\documentclass[10pt,journal]{IEEEtran}

\usepackage[T1]{fontenc}
\usepackage[utf8]{inputenc}
\usepackage{amsmath,amssymb,amsfonts}
\usepackage{graphicx}
\usepackage{booktabs}
\usepackage{multirow}
\usepackage{array}
\usepackage{tikz}
\usetikzlibrary{arrows.meta,positioning,fit,backgrounds,shapes.geometric,calc}
\usepackage{pgfplots}
\pgfplotsset{compat=1.17}
\usepackage{caption}
\usepackage{subcaption}
\usepackage{xcolor}
\usepackage{enumitem}
\usepackage[colorlinks=true,citecolor=blue,linkcolor=blue,urlcolor=blue]{hyperref}
\usepackage[capitalize]{cleveref}

\newcommand{\dacc}{\ensuremath{\mathrm{DAcc}}}
\newcommand{\oacc}{\ensuremath{\mathrm{OAcc}}}

\newcommand{\macrof}{\ensuremath{\mathrm{Macro\text{-}F1}}}

\title{AdapterMoE: A Two-Stage Hard-Routing Mixture-of-Experts Architecture for Multi-Crop Disease Recognition with Calibrated Rejection and Incremental Learning}

\author{Pin-Hsun~Huang and Shaou-Gang~Miaou%
\thanks{The authors are with the Department of Electronic Engineering, Chung Yuan Christian University, Taoyuan, Taiwan. Corresponding author: Shaou-Gang Miaou (e-mail: miaou@cycu.edu.tw).}}

\begin{document}
\raggedbottom

\maketitle

\begin{abstract}
Timely identification of crop diseases is critical to food security, and multi-crop recognition has a natural divide-and-conquer structure -- a good fit for Mixture-of-Experts (MoE). Conventional soft-routing MoE, however, learns this assignment freely through end-to-end training, which tends to let a few experts dominate (\emph{expert collapse}) with no semantic correspondence to the crops, and such models further face high retraining costs for new crops, unstable rejection of non-target inputs, and an accuracy ceiling largely saturated across competing methods. We therefore shift the objective from maximizing accuracy toward a trade-off among deployment cost, scaling flexibility, and rejection stability, and adopt deterministic hard routing instead. We propose \textbf{AdapterMoE}, a two-stage hard-routing architecture: a RouterHead first classifies the crop and rejects non-target crops via a Maximum Softmax Probability (MSP) threshold, with a dual-gate out-of-distribution (OOD) module (Energy + $K$-nearest-neighbor) intercepting distribution-shifted inputs; five per-crop Adapters atop a \emph{frozen} EfficientNet-B0 backbone then discriminate among diseases, each independently calibrated via Temperature Scaling. Because experts are hard-isolated at the data level, the architecture avoids expert collapse by design and exposes an \texttt{add\_crop} interface, whereby adding a crop is a local operation rather than a full retrain. On PlantVillage (5 crops, 26 classes), across a fair five-system comparison under five random seeds, AdapterMoE attains accuracy statistically indistinguishable from the best competing systems (Macro-F1 within a 0.24-point band) while cutting training cost to $\sim$9\% of the full-network baselines, expanding to a new crop in $1.14$ minutes on average, and exhibiting the lowest cross-seed rejection-accuracy variance of all five systems. A zero-shot evaluation on the field-condition PlantDoc dataset shows the dual-gate module rejects true out-of-distribution crops at $95.7\pm2.8\%$, while in-domain classification degrades sharply under covariate shift -- a limitation we attribute honestly to dataset shift rather than to the routing design. AdapterMoE's contribution is therefore not a state-of-the-art accuracy claim but a reproducible, transparently reported trade-off point for deployment cost, extensibility, and rejection reliability.
\end{abstract}


\section{Introduction}
\label{sec:intro}

\IEEEPARstart{C}{ontinued} population growth and accelerating climate change have made food security one of the most closely watched issues in modern agriculture, and plant disease remains one of the principal threats to crop yield. A global synthesis by Savary et al.~\cite{savary2019} estimates that pathogens and pests reduce the yield of the world's major staple crops -- wheat, rice, maize, potato, and soybean -- by 17.2\% to 30.0\% annually, a scale of loss with direct consequences for food supply and national agricultural economies. Diagnosis has traditionally relied on farmers' field experience, which is slow and whose quality depends on individual knowledge and observation conditions; the pressures of expanding cultivated area and an aging agricultural workforce make the limits of manual diagnosis increasingly apparent.

Deep-learning-based image recognition has opened new possibilities for automated diagnosis. Hughes and Salath\'e~\cite{hughes2015plantvillage} released the PlantVillage dataset in 2015 -- roughly 54k labeled images spanning 14 crops and 38 disease-or-healthy classes captured under controlled conditions -- which quickly became the field's most widely used benchmark, with subsequent CNN-based work routinely exceeding 99\% accuracy on it. Ferentinos~\cite{ferentinos2018}, however, showed in 2018 that models trained only on such laboratory-condition imagery degrade markedly on real field photographs, exposing a persistent gap between \emph{controlled benchmarks} and \emph{field deployment}. Moving existing methods from the lab to practice therefore still requires confronting data diversity, rejection capability, and interpretability.

In practice, a user-submitted leaf photograph may span several crops and may also contain inputs the training data never covered -- new cultivars, background weeds, or simply a photo that is not a leaf at all. This creates a dual requirement: \emph{multi-crop recognition} and \emph{rejection of non-target inputs}. Most existing work, however, targets fine-grained classification within a single crop and rarely treats cross-crop feature interference and open-set inputs systematically. This work asks how to design a single end-to-end system that satisfies both multi-crop classification accuracy and reliable rejection of non-target inputs at once.

\subsection{Motivation and Problem Statement}
\label{sec:motivation}

Closer inspection of existing approaches surfaces four core, mutually reinforcing challenges.

\textbf{(1) Cross-crop feature interference.} A single flat classifier trained directly over all 26 classes must learn discriminative boundaries for 5 crops and 25 leaf classes within one shared feature space, even though different crops differ substantially in leaf shape, texture, and color (a grape leaf and a tomato leaf look very different), while diseases \emph{within} the same crop can look nearly identical. This ``large inter-crop, small intra-crop'' heterogeneity forces a single feature extractor to compromise between distinguishing crops and discriminating fine-grained disease, and the resulting monolithic feature space is difficult to maintain: adding a crop touches the shared representation and forces full retraining, with no structured pathway for rejecting non-target crops. We frame multi-crop disease recognition as a ``triage, then diagnose'' pipeline: the system first identifies which crop an image belongs to, then hands it to that crop's own specialist for fine-grained diagnosis.

\textbf{(2) The need to reject an ``Others'' category.} Deployed inputs may be a non-target crop, a novel disease absent from the label set, or not a leaf at all. A system forced to classify over a fixed 25-class label set will produce confident, wrong predictions for such inputs -- with real consequences in agriculture, such as misapplied pesticide. Because most prior evaluations use closed test sets only, this failure mode is rarely exposed. We treat Others-rejection as a first-class objective, quantified by a dedicated Others Accuracy (\oacc) metric.

\textbf{(3) Structural collapse of soft-routing Mixture-of-Experts (MoE).} MoE is a natural fit for cross-crop heterogeneity -- different experts can specialize in different crops -- but Shazeer et al.~\cite{shazeer2017} document \emph{expert collapse} in end-to-end soft-routing MoE: a few experts, selected more often early in training, are reinforced and come to dominate while the rest atrophy. We additionally observe that even with an auxiliary load-balancing loss, gate weights remain markedly uneven (Section~\ref{sec:ablation-lb}); this structural failure mode is difficult to remove through auxiliary losses alone once it is baked into the architecture.

\textbf{(4) A systemic lack of uncertainty quantification.} Rejection thresholds in prior work are mostly set empirically. Maximum Softmax Probability (MSP)~\cite{hendrycks2017} is widely used, but modern deep networks are systematically overconfident~\cite{guo2017}, so raw MSP deviates from the true posterior and a threshold on it lacks a clean statistical interpretation.

These four issues are mutually entangled: pursuing accuracy alone cannot solve rejection; adding a rejection mechanism without probability calibration yields an unreliable threshold; and adopting an MoE architecture without addressing collapse can make the system \emph{more} brittle, not less. This work proposes a single integrated architecture that addresses all four simultaneously.

\subsection{Contributions}
\label{sec:contributions}
We pursue an integrated architecture addressing all four challenges above, with contributions at four levels:
\begin{enumerate}[leftmargin=*,itemsep=2pt]
\item \textbf{Architecture.} AdapterMoE: a two-stage hard-routing architecture built from a shared frozen EfficientNet-B0 backbone, a RouterHead, five per-crop Adapters, and a DirectHead, with dual-gate routing (MSP + Energy + KNN) directing each input down one of four decision paths (High-confidence, Low-confidence fusion, Others-reject, or out-of-distribution (OOD)-reject). Deterministic routing eliminates soft-routing's expert collapse by design (Challenge 3) while an independently calibrated Temperature Scaling per branch gives the MSP threshold and dual-gate OOD mechanism a genuine probabilistic reading (Challenges 2 and 4). Backbone freezing with feature caching also lets the main pipeline train at a small fraction of the other systems' cost -- though, as made explicit in Section~\ref{sec:cost}, this stems from the frozen-backbone-plus-caching training paradigm and is orthogonal to the hard-vs-soft routing question itself. The dual-gate OOD mechanism's benefit under in-distribution-only evaluation is itself small and seed-dependent (Section~\ref{sec:ablation-ood}), so we report it as a conditional, deployment-dependent contribution rather than an unconditional improvement.
\item \textbf{Incremental-learning interface.} The \texttt{add\_crop} interface (Challenge 1) lets a new crop be added by training one new Adapter and lightly fine-tuning the Router and DirectHead, leaving all other crop modules untouched; existing crops' rejection accuracy remains stable throughout expansion (Section~\ref{sec:addcrop}). No other system in our comparison offers this capability.
\item \textbf{Rejection stability.} Across five random seeds, evaluated in a fair comparison framework against three flat baselines and one soft-routing MoE baseline (Section~\ref{sec:fair-comparison}), AdapterMoE attains the highest mean Others-Accuracy and the lowest cross-seed standard deviation of all five systems -- a property directly relevant to writing a deployment service-level agreement.
\item \textbf{Transparency.} On this task, accuracy has saturated: all four strong systems' Macro-F1 scores fall within a 0.24-point band, and we do not claim a state-of-the-art accuracy result. We instead report low-cost extensibility as the most robust contribution, and disclose every negative or conditional finding -- including that AdapterMoE's Macro-F1 is marginally below the soft-routing baseline's (within that baseline's own cross-seed standard deviation) -- alongside the evidence for it.
\end{enumerate}

The remainder of the paper is organized as follows. Section~\ref{sec:related} surveys related work in crop disease recognition, Mixture-of-Experts, and rejection/uncertainty quantification. Section~\ref{sec:method} details the AdapterMoE architecture. Section~\ref{sec:experiments} reports the five-system comparison, incremental learning, ablations, and an external zero-shot evaluation on PlantDoc. Section~\ref{sec:discussion} discusses limitations, and Section~\ref{sec:conclusion} concludes with future directions.

\section{Related Work}
\label{sec:related}

\subsection{Deep Learning for Crop Disease Recognition}
\label{sec:related-cnn}

\textbf{Datasets and CNN backbones.} PlantVillage~\cite{hughes2015plantvillage} established the field's most-used benchmark, and early CNN work regularly exceeded 99\% accuracy on it. Ferentinos~\cite{ferentinos2018} showed that this controlled-condition accuracy does not transfer to field photographs, tracing the gap to \emph{covariate shift} at the dataset level: PlantVillage images have clean, uniform backgrounds and a single centered leaf, while field images have clutter, variable lighting, occlusion, and inconsistent framing -- a dataset-level mismatch rather than an architectural flaw. We retain PlantVillage as our primary set for comparability with prior work, while probing this exact limitation in Section~\ref{sec:plantdoc}. For backbones, He et al.'s ResNet~\cite{he2016resnet} (skip connections; ResNet-18, 11.2M parameters) has long served as a standard leaf-classification baseline, and Simonyan and Zisserman's VGG~\cite{simonyan2015vgg} was an early but less parameter-efficient alternative. Tan and Le's EfficientNet~\cite{tan2019efficientnet} scales depth, width, and resolution jointly via a compound coefficient $\varphi$; EfficientNet-B0 (4.04M parameters) reaches higher ImageNet accuracy than ResNet-18 at roughly a third of the parameters, making it a common recent choice for this task, and we adopt it as our shared backbone after comparing both in preliminary trials.

\textbf{Vision Transformers.} Dosovitskiy et al.'s ViT~\cite{dosovitskiy2021vit} treats an image as a sequence of patch tokens processed by self-attention, triggering a broader shift in computer vision backbones; Liu et al.'s Swin Transformer~\cite{liu2021swin} adds hierarchical shifted-window attention, and Touvron et al.'s DeiT~\cite{touvron2021deit} uses distillation to reduce ViT's dependence on large-scale pretraining. We nonetheless adopt a CNN backbone rather than a ViT for four reasons. First, PlantVillage's $\sim$54k images are a moderate-scale dataset relative to ImageNet, and ViT's generalization at this scale depends more heavily on large-scale pretraining than a CNN's does. Second, our central contribution is the routing and rejection mechanism, not the backbone itself, so a mature, stable CNN backbone lets us isolate the methodological contribution. Third, ViT's inference latency and memory footprint exceed a comparable CNN's, working against future edge deployment. Fourth -- and most directly supported by our own results -- Section~\ref{sec:main-results} shows that the CNN backbones we do compare (SimpleCNN, ResNet-18, EfficientNet-B0) already land within one standard deviation of one another in Macro-F1, indicating that backbone representational capacity is not the current bottleneck on this task; introducing a more complex, more data-hungry ViT at this point would be a poor allocation of modeling effort. We therefore leave ViT-family backbones (including Swin and DeiT) as future integration work.

\textbf{Classification versus detection.} Some prior work frames crop disease recognition as object detection -- localizing a leaf or lesion before classifying the crop within it -- rather than as whole-image classification. We frame it as classification for three reasons. First, PlantVillage images are single-leaf, pre-cropped, and centered, with no bounding-box annotation; building a detection-ready dataset is outside our current scope. Second, our target deployment scenario is a user photographing a single leaf at close range (e.g., a phone-app diagnosis feature), where detection's core problem -- localizing multiple targets in a cluttered scene -- is not the binding constraint; detection's advantage shows up specifically when a raw field photo contains multiple plants, multiple leaves, and background clutter, which is outside our current data and application scope. Third, the two framings are not mutually exclusive: detection can be layered in front of our classifier as a pre-processing stage that crops a single leaf out of a cluttered raw photo before handing it to our recognition system, which we list as future work (Section~\ref{sec:future}).

\subsection{Mixture-of-Experts}
\label{sec:related-moe}

The core idea of Mixture-of-Experts is divide-and-conquer: a complex input space is split into subtasks, each handled by its own expert network, with a gate or router deciding each expert's participation. Jacobs et al.~\cite{jacobs1991moe} laid the theoretical foundation in 1991 with \emph{Adaptive Mixtures of Local Experts}: a gating network produces per-expert weights from the input, and the final output is their weighted combination, trained end-to-end so that experts differentiate naturally through competition for data. The key insight -- that not every input need be processed by the same network -- is especially apt for heterogeneous tasks such as multi-crop, multi-disease classification, where a single network must otherwise compromise across sub-distributions. This original design used \emph{soft} (dense) gating, however, so every expert executes a forward pass for every input, and compute grows linearly with the number of experts.

Shazeer et al.~\cite{shazeer2017} brought MoE to the large-model era via sparsely-gated Top-$K$ routing, activating only the $K$ (typically 1--2) highest-weighted experts per input and decoupling model capacity from per-input compute -- but this introduced expert collapse: experts selected more often early in training receive more gradient signal, become more competitive, and are selected even more, starving the rest, which Shazeer et al. mitigate with an auxiliary load-balancing loss. Follow-ups simplify or soften this routing: Fedus et al.'s Switch Transformer~\cite{fedus2022switch} uses Top-1 routing at trillion-parameter scale, and Puigcerver et al.'s Soft MoE~\cite{puigcerver2023softmoe} replaces the discrete Top-$K$ operation with soft per-token weighting. In vision, Riquelme et al.'s V-MoE~\cite{riquelme2021vmoe} integrated sparse gating into ViT at ImageNet scale; Mu and Lin's 2025 survey~\cite{mu2025moesurvey} finds expert collapse and load imbalance remain the central open challenges for soft-routing MoE across vision, language, and multimodal settings.

Applying this line of work directly to a medium-scale, strongly heterogeneous task such as multi-crop disease recognition runs into two structural issues. First, a load-balancing loss can mitigate collapse at the level of expert utilization but does not eliminate it: our own controlled ablation (Section~\ref{sec:ablation-lb}) shows that even with the loss enabled, the soft-routing baseline still exhibits a maximum gate probability of $\mathrm{MaxGate}\approx0.35$ across five seeds -- and disabling the loss drives $\mathrm{MaxGate}$ to $0.81$, a clear single-expert-dominance regime. Second, conventional MoE designs presume expert specialization should emerge from end-to-end data competition rather than from a semantic prior; when the input space has an explicit semantic hierarchy (crop, then disease), letting experts differentiate freely discards the interpretable correspondence to actual crops. These two limitations directly motivate our shift to two-stage hard routing (Section~\ref{sec:method}).

\subsection{Rejection and Uncertainty Quantification}
\label{sec:related-rejection}

Closed-set classification assumes every test sample belongs to a known class, but deployed systems routinely see inputs that do not -- a non-target crop, or a disease absent from the training label set. A rejection mechanism lets the model say ``I should not answer this'' instead of emitting a confident wrong prediction. Hendrycks and Gimpel~\cite{hendrycks2017} proposed Maximum Softmax Probability (MSP) as a simple baseline: correctly classified in-distribution samples tend to have higher MSP than misclassified or OOD ones, so a single scalar threshold suffices for basic rejection without modifying or retraining the model. MSP has two known weaknesses, however: modern deep networks are systematically overconfident, so MSP can be high even for wrong predictions, and MSP has no inherent probabilistic meaning -- an MSP of 0.7 does not mean a true 70\% posterior -- so a threshold on it lacks statistical grounding.

Guo et al.~\cite{guo2017} used Expected Calibration Error (ECE)~\cite{naeini2015ece} to show systematically that modern deep networks such as ResNet and DenseNet, despite high top-1 accuracy, are markedly overconfident, and proposed \emph{Temperature Scaling}: dividing the logit vector by a single scalar temperature $T$ before the softmax, fit on a held-out calibration split by minimizing negative log-likelihood. Temperature Scaling leaves the model's arg-max prediction unchanged, requires fitting only one parameter (converging in seconds), and typically drives ECE from several percentage points to near zero -- performance competitive with more complex calibrators such as Platt scaling or isotonic regression. We independently calibrate the Router, each Expert, and the DirectHead, which is necessary precisely because they output different numbers of classes (6, 3--10, and 26 respectively) and are therefore not comparable on a raw-logit scale.

Beyond softmax-based rejection, Liu et al.'s Energy score~\cite{liu2020energy} uses the full logit vector rather than only the top-1 probability, and Sun et al.'s deep-KNN detector~\cite{sun2022knn} instead measures distance in feature space to a bank of training-set embeddings; both target OOD inputs that MSP alone may miss. We build on both: an Energy score and a KNN-distance signal, combined by \texttt{AND} logic, form the second gate in our dual-gate rejection module (Section~\ref{sec:router}).

\subsection{Positioning}
\label{sec:related-positioning}
Our work sits at the intersection of three lines above. On the application spectrum, we target a multi-crop, multi-disease, Others-aware task rather than pushing marginal backbone accuracy within a single crop -- and, following Ferentinos~\cite{ferentinos2018}, we treat closed-set accuracy as necessary but not sufficient for deployment quality. On the MoE spectrum, we replace end-to-end soft gating with a two-stage hard-routing design that eliminates expert collapse by construction rather than mitigating it after the fact, at the cost of requiring an explicit semantic hierarchy in the task (which multi-crop disease recognition naturally provides). On the rejection/uncertainty spectrum, we integrate point-estimate MSP rejection with Temperature-Scaling-based calibration and an auxiliary Energy+KNN gate within a single pipeline -- a combination we find little precedent for in the crop-disease literature, and which we view as the paper's most specific positioning: bridging the structural task-decomposition literature with the calibrated-uncertainty-quantification literature.

\section{Method}
\label{sec:method}

\subsection{Overview}
\label{sec:overview}

\begin{figure*}[t]
\centering
\resizebox{0.92\linewidth}{!}{\begin{tikzpicture}[
    font=\scriptsize,
    box/.style={draw, rounded corners=2pt, minimum height=0.7cm, align=center, line width=0.6pt, inner sep=3.5pt},
    stagebox/.style={draw, dashed, rounded corners=4pt, line width=0.7pt, inner sep=0pt},
    arr/.style={-{Latex[length=1.8mm]}, line width=0.6pt},
    lbl/.style={font=\scriptsize, fill=white, inner sep=1pt}
]

\node[box, fill=blue!8, minimum width=2.6cm] (input) at (-1.0, 7.1) {Input leaf image\\$224{\times}224{\times}3$};
\node[box, fill=black!5, minimum width=6.6cm] (backbone) at (4.6, 7.1) {Shared \textbf{frozen} EfficientNet-B0 backbone (4.04M params)};
\draw[arr] (input) -- (backbone.west);

\node[box, fill=red!15, minimum width=2.4cm] (path0) at (0, 5.3) {\textbf{Path 0}: Others\\direct-reject};
\node[box, fill=blue!15, minimum width=3.0cm] (router) at (5.4, 5.3) {RouterHead\\6-way (5 crops + Others)};
\node[box, fill=red!15, minimum width=2.4cm] (path3) at (0, 4.15) {\textbf{Path 3}: OOD\\reject};
\node[box, fill=green!12, minimum width=3.0cm] (ood) at (5.4, 4.15) {Dual-gate OOD\\(Energy + KNN)};

\node[stagebox, fit=(path0)(router)(path3)(ood), inner ysep=20pt, inner xsep=6pt] (stage1) {};
\node[font=\scriptsize\bfseries, anchor=north west] at ($(stage1.north west)+(0.15,-8pt)$) {Stage 1};

\draw[arr] (backbone.south) -- (router.north);
\draw[arr] (router) -- node[lbl,pos=0.5,above=1pt]{arg-max$=$Others} (path0);
\draw[arr] (router) -- node[lbl,pos=0.5,right=1pt]{not Others} (ood);
\draw[arr] (ood) -- node[lbl,pos=0.5,above=1pt]{both fire} (path3);

\node[box, fill=orange!15, minimum width=3.2cm] (adapter) at (1.6, 1.7) {Per-crop Adapter\\(1-of-5 selected)};
\node[box, fill=green!12, minimum width=3.2cm] (directhead) at (6.4, 1.7) {DirectHead\\safety net (26-way)};
\node[box, fill=green!8, minimum width=4.4cm] (conf) at (4.0, 0.7) {Confidence router (calibrated MSP)};
\node[box, fill=blue!15, minimum width=2.8cm] (path1) at (1.6, -0.35) {\textbf{Path 1} (High):\\Adapter only};
\node[box, fill=orange!15, minimum width=3.4cm] (path2) at (6.4, -0.35) {\textbf{Path 2} (Low):\\Adapter $\oplus$ DirectHead fusion};

\node[stagebox, fit=(adapter)(directhead)(conf)(path1)(path2), inner ysep=20pt, inner xsep=6pt] (stage2) {};
\node[font=\scriptsize\bfseries, anchor=north west] at ($(stage2.north west)+(0.15,-8pt)$) {Stage 2};

\draw[arr] (ood.south) -- node[lbl,pos=0.5,right=1pt]{passed (target crop)} ++(0,-0.45) -| (directhead.north);
\draw[arr] ($(ood.south)+(0,-0.45)$) -| (adapter.north);
\draw[arr] (adapter.south) -| ($(conf.north)+(-1.2,0)$);
\draw[arr] (directhead.south) -| ($(conf.north)+(1.2,0)$);
\draw[arr] (conf) -- node[lbl,pos=0.5,left=1pt]{high} (path1);
\draw[arr] (conf) -- node[lbl,pos=0.5,right=1pt]{low} (path2);

\node[box, fill=blue!10, minimum width=6.6cm] (output) at (4.0, -2.1) {Output: 25 disease classes (incl.\ healthy), or \textbf{Others}};
\draw[arr] (path0.west) -- ++(-0.8,0) |- ($(output.west)+(-2pt,0)$);
\draw[arr] (path3.west) -- ++(-1.6,0) |- ($(output.west)+(-2pt,0.2)$);
\draw[arr] (path1.south) -| ($(output.north)+(-2.4,2pt)$);
\draw[arr] (path2.south) -| ($(output.north)+(2.4,2pt)$);

\node[box, fill=purple!10, text width=2.6cm, minimum height=0.55cm, font=\tiny] (ts) at (2.1, -3.4) {\textbf{Temperature Scaling}: calibrates all 7 heads};
\node[box, fill=orange!12, text width=3.2cm, minimum height=0.55cm, font=\tiny] (addcrop) at (5.9, -3.4) {\textbf{\texttt{add\_crop}}: incremental interface, existing weights frozen};
\node[font=\tiny\itshape, anchor=east] (spanslbl) at ($(ts.west)+(-0.1,0)$) {Spans whole pipeline};
\node[stagebox, fit=(ts)(addcrop)(spanslbl), inner ysep=8pt, inner xsep=8pt] (mech) {};

\end{tikzpicture}}
\caption{AdapterMoE overall architecture. The shared frozen backbone feeds RouterHead, which first separates Others from the five target crops (Path 0 if Others) and, for samples it assigns to a crop, passes them to a dual-gate Energy-\textbf{AND}-KNN check that screens for distribution shift (Path 3 if both fire); samples surviving both checks then reach the selected crop's Adapter and the DirectHead safety net, whose outputs a calibrated-MSP confidence router combines into a high-confidence path (Adapter only) or a low-confidence path (log-odds fusion) on the way to the final prediction. Temperature Scaling and the \texttt{add\_crop} incremental interface (bottom) apply across this entire pipeline rather than to any single stage (Sections~\ref{sec:calibration} and~\ref{sec:addcrop}).}
\label{fig:architecture}
\end{figure*}
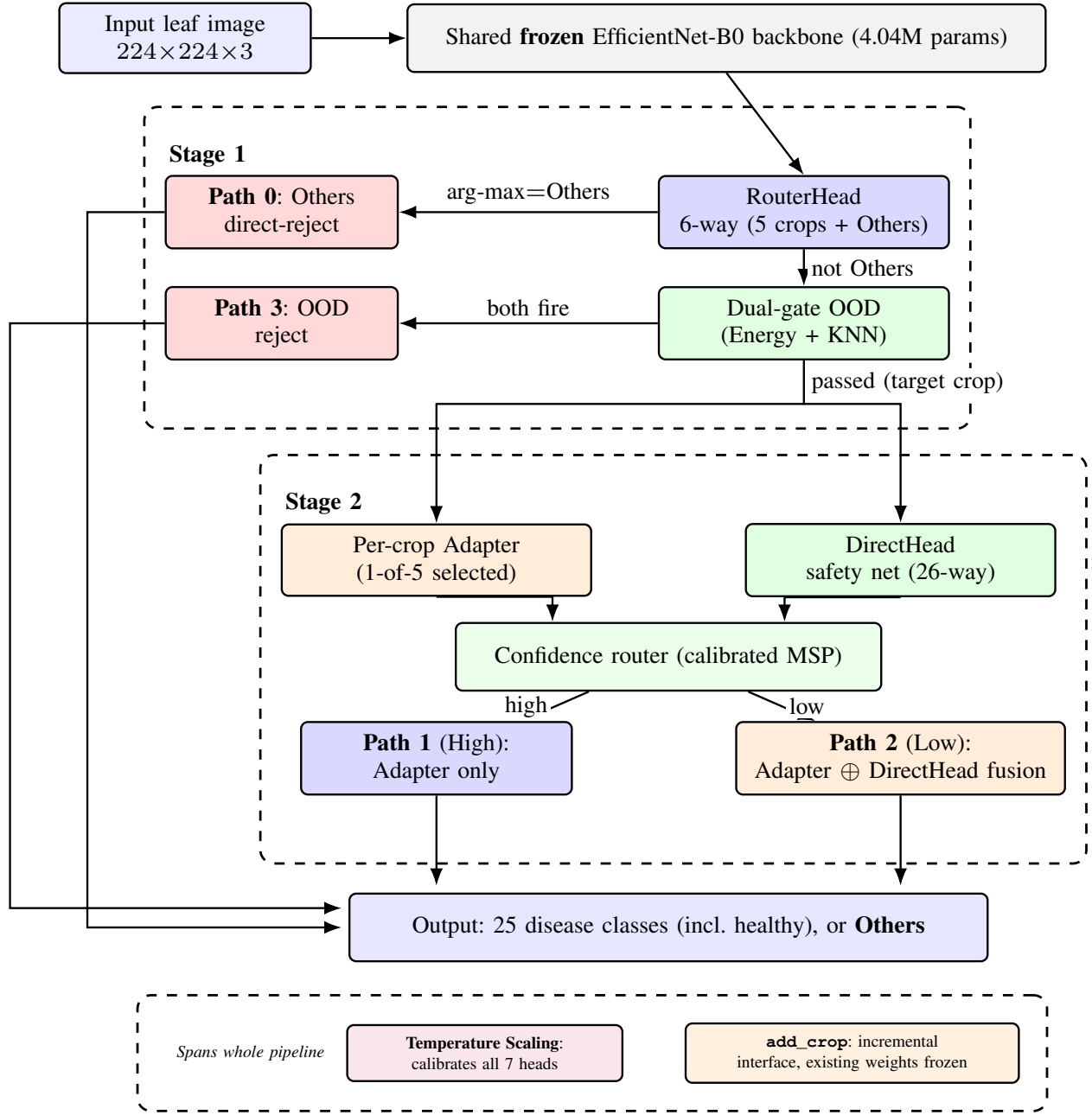

AdapterMoE is a two-stage hard-routing architecture with six functional modules. Given an input leaf image, the pipeline runs: feature extraction $\rightarrow$ crop routing with dual-gate OOD detection $\rightarrow$ per-branch Temperature Scaling $\rightarrow$ disease classification, terminating in one of four mutually exclusive decision paths (Others-reject, high-confidence Adapter-only, low-confidence log-odds fusion, or OOD-reject). \Cref{fig:architecture} shows the full pipeline; \Cref{fig:addcrop} illustrates the incremental-learning interface described in Section~\ref{sec:addcrop}.

The design can be read by analogy to hospital triage: the RouterHead is the intake desk routing a patient to the right specialist or turning them away (Others); the five per-crop Adapters are specialists who only diagnose within their own crop; and the DirectHead is an emergency room providing backup when the Router is unsure or an Adapter's judgment sits near a decision boundary. This yields two benefits: adding a crop is like hiring one specialist rather than restaffing the hospital (\texttt{add\_crop}, Section~\ref{sec:addcrop}), and since Adapters share no gradients, the architecture is immune by construction to soft-routing's expert collapse.

\textbf{Stage 1 (Router).} The system first resizes the input to $224\times224$ RGB and extracts a 1280-d feature $h(x)$ from a shared, frozen EfficientNet-B0 backbone. The RouterHead is a 6-way classifier over \{Apple, Corn, Grape, Potato, Tomato, Others\}. If the arg-max class is Others, the sample is rejected immediately (Path 0) and the pipeline terminates.

\textbf{Stage 2 (Expert).} If the Router selects one of the five target crops, the sample is handed to that crop's Adapter, which classifies only among that crop's own disease classes (e.g., 4 for Apple, 10 for Tomato). Because each Adapter is trained solely on its own crop's sub-dataset, its feature space is far simpler than a flat 26-way classifier's -- the central source of hard routing's advantage over a monolithic model.

\textbf{Path selection.} Before reaching the Adapter, non-Others samples pass through a dual-gate OOD check (Section~\ref{sec:router}); survivors are then split by the Router's calibrated MSP into a high-confidence zone ($\mathrm{MSP}\geq0.55$, Path 1: Adapter output used directly) and a low-confidence zone ($\mathrm{MSP}<0.55$, Path 2: Adapter and DirectHead logits combined by fixed-weight log-odds fusion, Section~\ref{sec:fusion}). This two-zone design applies a cheaper decision rule to confident samples and a more conservative, multi-branch one to uncertain samples.

\textbf{DirectHead safety net.} An independent 26-way flat classifier, DirectHead, is trained on the same flat 26-class configuration used to evaluate all five systems (Section~\ref{sec:fair-comparison}). It is not a primary classifier but a fallback: architecturally, it is the same family as the flat EfficientNet-B0 baseline, but here it functions only as one input to the low-confidence fusion rule, giving the system a "universal backup" so that a Router misjudgment at a decision boundary does not cascade into total failure.

\textbf{Calibration layer.} Because the Router (6-way), each Expert (3--10-way), and DirectHead (26-way) output different numbers of classes, their raw logits are not directly comparable. Each of the seven branches (Router, five Adapters, DirectHead) is independently fit with a scalar Temperature Scaling parameter (Section~\ref{sec:calibration}), after which MSP carries a genuine probabilistic meaning and the two-zone threshold and log-odds fusion are built on a consistent probability scale.

\textbf{Parameter budget.} The full AdapterMoE pipeline uses a single ImageNet-pretrained, frozen EfficientNet-B0 backbone with seven lightweight heads (one RouterHead, five per-crop Adapters, one DirectHead) attached, for a total of 11.31M parameters -- essentially the same order as the flat EfficientNet-B0 baseline (4.04M) and the soft-routing baseline (11.25M), enabling a like-for-like comparison (Section~\ref{sec:main-results}).

\subsection{Dataset and Preprocessing}
\label{sec:dataset}

We use PlantVillage~\cite{hughes2015plantvillage}, selecting five crops -- Apple, Corn, Grape, Potato, Tomato -- comprising 25 leaf classes (including each crop's healthy class), plus an ``Others'' class built from PlantVillage images of crops outside this set (e.g., pepper, strawberry, cherry), for 26 classes total. Class counts are imbalanced: Tomato has the most sub-classes (10) and Potato the fewest (3), making the Tomato Adapter the system's heaviest-loaded sub-classifier (see the ablation in Section~\ref{sec:ablation-tomato}).

We derive three data configurations from this source, matched to each module's training need: (i) a \emph{crop-level} configuration (6 classes: 5 crops + Others) for the Router, which merges all diseases within a crop into one class and explicitly decouples crop identification from disease diagnosis at the data level; (ii) a \emph{flat 26-class} configuration, shared across all five systems for training the flat baselines and ClassicalMoE\_Shared and for unified evaluation (also DirectHead's own training set); and (iii) five \emph{per-crop} sub-datasets for Adapter training, each containing only that crop's 3--10 disease classes.

All three configurations are split into Train/Val/Test at a fixed ratio. Following the independent-calibration-set practice of Guo et al.~\cite{guo2017}, we further split each Val set 1:1 into \texttt{val\_a} (used only for loss monitoring, not for any decision) and \texttt{val\_b} (used exclusively, after training is complete, to fit each branch's Temperature Scaling parameter, with no gradient or early-stopping role) -- and this split is performed independently for the Router, each Expert, and DirectHead so that no branch's calibration data is shared with another's.

All images are resized to $224\times224$ and normalized with ImageNet statistics ($\mu=(0.485,0.456,0.406)$, $\sigma=(0.229,0.224,0.225)$). Training-time augmentation uses two intensities: a conservative transform for the 25 named disease classes (random horizontal flip, $\pm15^\circ$ rotation, light color jitter of $\pm20\%$ on brightness/contrast/saturation) chosen to avoid destroying the fine lesion, curl, and discoloration textures that disease recognition depends on; and a stronger transform for Others (adding vertical flip, $\pm30^\circ$ rotation, $\pm30\%$ color jitter, $\pm0.1$ hue jitter, and 30\% probability random erasing) intended to widen Others' visual diversity so the Router learns a general ``does not resemble any target crop'' boundary rather than memorizing the specific Others training images. We apply the same Others-augmentation policy to \emph{all five} systems in the comparison for fairness.

We repeat every experiment under five random seeds $\{42,123,456,789,2026\}$, use deterministic computation throughout (at a measured 5--10\% training-speed cost), and report all headline metrics as mean $\pm$ sample standard deviation (obtained with an unbiased estimate of sample variance).

\subsection{Stage 1: Router and Others Rejection}
\label{sec:router}

The RouterHead is a lightweight MLP over the shared 1280-d feature: $\mathrm{Linear}(1280{\to}256) \to \mathrm{LayerNorm} \to \mathrm{ReLU} \to \mathrm{Dropout}(0.1) \to \mathrm{Linear}(256{\to}6)$, roughly 0.33M parameters, with no backbone of its own. It is trained with Adam at a lower initial learning rate ($5\times10^{-4}$ versus $10^{-3}$ elsewhere) and label smoothing of 0.05 (not used elsewhere), both chosen to reduce cross-seed variance in Router quality, for 15 epochs (matching DirectHead).

Rejection uses a two-layer rule on top of the calibrated softmax output. First, if the arg-max class is Others, the sample is rejected immediately (Path 0), at minimal computational cost. Second, for samples whose arg-max is one of the five crops, we additionally require $\mathrm{MSP}\geq\theta$ (we use $\theta=0.55$) to trust that assignment with high confidence; below threshold, the sample is instead routed to the low-confidence fusion path (Section~\ref{sec:fusion}). Threshold sensitivity is examined directly in Section~\ref{sec:ablation-theta}.

As a complementary OOD signal, we add an Energy score~\cite{liu2020energy}, defined for a logit vector $z\in\mathbb{R}^K$ and temperature $\tau$ as
\begin{equation}
E(x;f) = -\tau\log\left(\sum_{i=1}^{K}\exp(z_i/\tau)\right),
\label{eq:energy-general}
\end{equation}
which we use at $\tau=1$ following the original authors' recommendation that larger $\tau$ flattens the score and reduces its discriminative power between in- and out-of-distribution samples, so that
\begin{equation}
E(x) = -\log\left(\sum_{i=1}^{K}\exp(z_i)\right).
\label{eq:energy}
\end{equation}
This $\tau$ is a fixed constant inside the Energy computation, unrelated to the learned Temperature Scaling parameter $T$ of Section~\ref{sec:calibration}. Where MSP uses only the top-1 probability, Energy uses the whole logit vector and typically takes larger values for inputs far from the training distribution. Rather than an arbitrary Energy threshold, we set it (and, analogously, a K-Nearest-Neighbor (KNN) distance threshold over an L2-normalized 1280-d feature bank with $k=50$, where $k$ denotes the number of nearest neighbors) from the in-distribution validation split at a target true-positive rate of 0.95. A sample is flagged OOD (Path 3, forced to Others) only when \emph{both} signals fire -- an \texttt{AND} rule chosen over \texttt{OR} or either alone in Section~\ref{sec:ablation-ood}. This gate is independent of the MSP threshold: MSP governs the high/low-confidence split \emph{within} accepted target-crop samples, while the OOD gate catches samples routed to a crop with high top-1 probability but a weak overall logit profile -- an overconfidence failure mode MSP alone can miss.

Because the raw $\theta=0.55$ threshold has no probabilistic meaning without calibration, we fit a Router-specific Temperature Scaling parameter on the \texttt{val\_b} split (Section~\ref{sec:calibration}); after calibration, ``$\mathrm{MSP}\geq0.55$'' means, to a good approximation, ``the model's posterior probability that this is a target crop exceeds 55\%,'' rather than simply an empirically convenient softmax cutoff.

\subsection{Stage 2: Per-Crop Adapters}
\label{sec:adapter}

Each Adapter (we use ``Expert'' and ``Adapter'' interchangeably) handles only its own crop's disease classes -- 4 for Apple, Corn, and Grape, 3 for Potato, 10 for Tomato -- and is trained only on that crop's own sub-dataset, never seeing another crop's images or the Others class. This is the key structural difference from soft routing: expert specialization here is enforced at the \emph{data} level, not left to emerge from gating dynamics during training.

The main pipeline uses a three-step lightweight design. \textbf{Step 1 (backbone pre-calibration and freezing):} an ImageNet-pretrained EfficientNet-B0 is briefly fine-tuned end-to-end for 3 epochs on the flat 26-class training configuration -- enough to adapt it somewhat to plant-leaf image statistics -- and then frozen entirely; all subsequent training operates on cached features rather than re-running the backbone. \textbf{Step 2 (lightweight heads on frozen features):} the RouterHead as above ($\sim$0.33M); a per-crop AdapterHead using a Squeeze-and-Excitation (SE) block~\cite{hu2018se} (reduction ratio 8) followed by $\mathrm{Linear}(1280{\to}512)\to\mathrm{ReLU}\to\mathrm{Dropout}(0.2)\to\mathrm{Linear}(512{\to}\text{\#classes})$, about 1.07M parameters each for Apple/Corn/Grape/Potato (\emph{StandardAdapter}); and, for the heavier 10-class Tomato task, an enhanced variant with SE reduction ratio 4, a two-layer MLP ($1024{\to}512$), LayerNorm, GELU, dropout 0.3, and label smoothing 0.05, at about 2.67M parameters (\emph{TomatoEnhancedAdapter}); plus a DirectHead as a single linear layer ($1280{\to}26$, $\sim$0.03M), cheap enough to train in minutes directly on cached features. The SE block rescales each feature channel by a learned, sigmoid-bounded importance weight, $x_{ch}' = s_{ch}\cdot x_{ch}$, where $x_{ch}$ is the original activation of channel $ch$ and $s_{ch}\in(0,1)$ is its learned per-channel gate, letting each Adapter learn which of the shared 1280 feature channels matter most for its own crop. \textbf{Step 3 (feature caching):} since the backbone is frozen, every training image's backbone feature is fixed across epochs, so it is computed once and cached; subsequent epochs read the cache directly. This yields roughly a $13\times$ training speedup for the Adapter stage relative to full backbone fine-tuning (the five Adapters together train in about 0.20 minutes; see Section~\ref{sec:cost} for the full-system cost comparison).

Adapters are trained with Adam ($\mathrm{lr}=10^{-3}$), Cosine Annealing, for a fixed 30 epochs with early stopping disabled (to avoid seed-dependent training-length variation), using unweighted cross-entropy (plus label smoothing 0.05 for the Tomato-enhanced variant only). Because each Adapter is trained fully independently, adding or replacing one crop's Adapter never requires touching another crop's weights -- the direct source of the modular-maintenance advantage exploited by \texttt{add\_crop} (Section~\ref{sec:addcrop}).

\subsection{DirectHead and Log-Odds Fusion}
\label{sec:fusion}

\subsubsection{Feature flattening}
An Adapter's output dimensionality equals its crop's disease-class count (3--10), but the system's final decision must live in the shared 26-class space. We therefore maintain a local-to-flat index map recording which flat-space index each of a crop's local classes corresponds to, and fill every flat-space position \emph{outside} that crop's class range with a large negative constant ($-1\times10^4$) as an ``impossible'' marker; after softmax, these positions carry negligible probability mass. This lets each Adapter ``speak only for its own classes'' -- e.g., the Apple Adapter explicitly abstains on the other 22 classes -- so that an Adapter's error on classes it was never trained on cannot systematically contaminate the fusion step below.

\subsubsection{Log-odds fusion}
The low-confidence path combines the calibrated, flattened Expert logit $z_e\in\mathbb{R}^{26}$ with the calibrated DirectHead logit $z_d\in\mathbb{R}^{26}$ by a fixed per-class linear combination:
\begin{equation}
z_{\text{fused}}[c] = \alpha \cdot z_e[c] + (1-\alpha)\cdot z_d[c], \qquad \forall c\in\{1,\dots,26\},
\label{eq:fusion}
\end{equation}
with the final prediction and probability given by
\begin{equation}
\hat{y} = \arg\max_c\, z_{\text{fused}}[c], \qquad
p_{\text{fused}}[c] = \frac{\exp(z_{\text{fused}}[c])}{\sum_{j=1}^{26}\exp(z_{\text{fused}}[j])}.
\label{eq:fusion-softmax}
\end{equation}
Because logits are (up to a translation) log-odds (Section~\ref{sec:calibration}), Equation~\ref{eq:fusion} is a log-odds-scale weighted combination of two branches' evidence, which is why calibrating $z_e$ and $z_d$ onto a common probability scale \emph{before} fusing them (Section~\ref{sec:calibration}) is a precondition for this combination to be statistically meaningful rather than an arbitrary heuristic.

We fix $\alpha=0.6$ rather than tuning it against validation accuracy, keeping fusion behavior reproducible across seeds and keeping any hard-vs-soft routing comparison uncontaminated by an extra tuned parameter. Because the low-confidence path (Path 2) triggers on only about 0.02\% of test samples in our PlantVillage setting (Section~\ref{sec:pathrouting}), $\alpha$'s influence on headline accuracy here is minimal; its value lies not in optimizing a fusion ratio but in providing a bounded-risk fallback for the rare case where Router and Adapter are simultaneously uncertain. We recommend a dedicated sensitivity analysis for deployments with more pronounced distribution shift, where this path is expected to trigger more often (Section~\ref{sec:ablation-gate}).

\subsubsection{Two-zone path selector}
The Router's calibrated MSP is a continuous confidence signal in $[0,1]$ rather than a binary trust decision, so the system splits it into two zones: a high-confidence zone (Path 1) where the Adapter output is trusted directly, and a low-confidence zone (Path 2) where Adapter and DirectHead are fused as above. This gives the routing decision a graceful-degradation property: when the Expert's confidence is insufficient, the system automatically broadens its evidence base rather than forcing a single mechanism to always be right. Combined with the dual OOD gate, the final decision resolves into four mutually exclusive paths:
\begin{itemize}[leftmargin=*,itemsep=1pt]
\item \textbf{Path 0 (Others-direct):} calibrated Router arg-max is Others; cheapest path, terminates immediately.
\item \textbf{Path 1 (high-confidence):} $\mathrm{MSP}\geq0.55$; the flattened Adapter arg-max is used directly, with no DirectHead computation.
\item \textbf{Path 2 (low-confidence fusion):} $\mathrm{MSP}<0.55$; Adapter and DirectHead are combined via Equation~\ref{eq:fusion}.
\item \textbf{Path 3 (OOD-reject):} the Energy and KNN gates both fire; output forced to Others.
\end{itemize}
This lets the system spend computation proportionally to a sample's difficulty: confident samples cost one Router pass plus one Adapter pass, while uncertain samples pay one additional DirectHead pass for extra robustness. Path usage rates and their conditional accuracies are reported in Section~\ref{sec:pathrouting}.

\subsection{Calibration Framework}
\label{sec:calibration}

For a $K$-class branch, let $z(x)\in\mathbb{R}^K$ be the pre-softmax logit vector and $y$ the ground-truth label. Temperature Scaling~\cite{guo2017} fits a single scalar $T>0$ per branch by minimizing Negative Log-Likelihood (NLL) on a held-out calibration split $D_{\text{calib}}$ (our \texttt{val\_b}):
\begin{equation}
T^{*} = \arg\min_{T>0} \; \mathcal{L}_{\mathrm{NLL}}(T) = -\frac{1}{N_{\text{cal}}}\sum_{i=1}^{N_{\text{cal}}} \log\,\mathrm{softmax}\!\left(\frac{z_i}{T}\right)\!\left[y_i\right],
\label{eq:nll}
\end{equation}
solved with L-BFGS (typically converging within 20--30 of a 200-step budget), with $T$ clamped to $[0.3,10.0]$ to avoid numerical instability from a poorly conditioned fit. We calibrate the Router, each of the five Adapters, and DirectHead independently on their own \texttt{val\_b} splits -- seven scalar temperatures in total, taking roughly 1--2 minutes on a single GPU as a one-time, offline step after training. This puts Router, Adapter, and DirectHead outputs on a common, trustworthy probability scale; gives the MSP thresholds $\theta_{\text{reject}}$ and $\theta_{\text{high}}$ an actual probabilistic reading rather than an arbitrary logit cutoff; and lets \texttt{add\_crop} (Section~\ref{sec:addcrop}) inherit a stable calibration baseline so newly added crops do not disturb existing crops' rejection behavior. Calibration quality (ECE before/after and the learned $T$ values) is reported in Section~\ref{sec:ablation-calibration}.

\subsection{The \texttt{add\_crop} Incremental-Learning Interface}
\label{sec:addcrop}

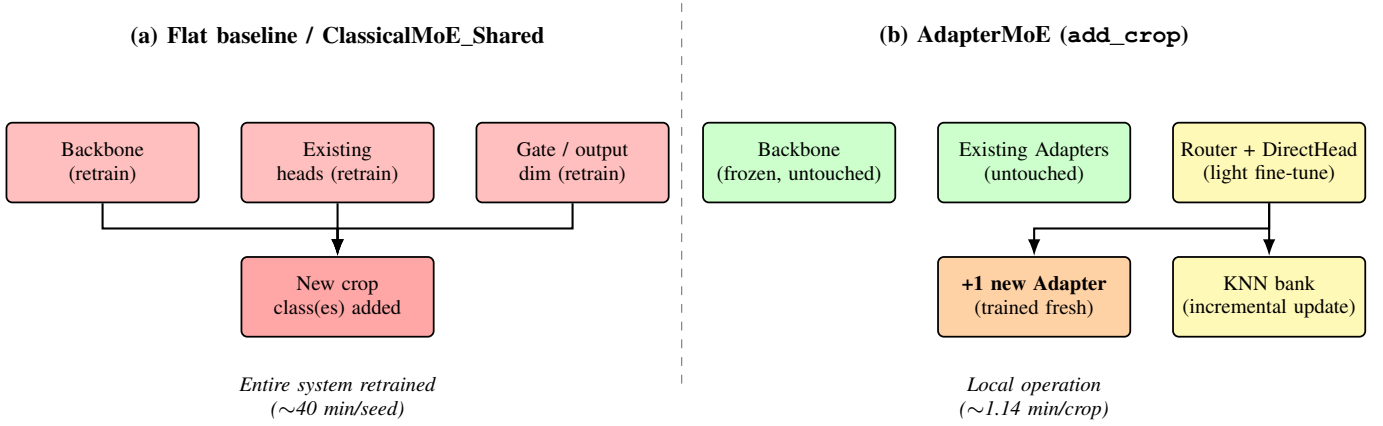
\begin{figure*}[t]
\centering
\resizebox{\linewidth}{!}{\begin{tikzpicture}[
    font=\scriptsize,
    box/.style={draw, rounded corners=2pt, minimum height=0.95cm, text width=2.15cm, align=center, line width=0.6pt, inner sep=2pt},
    arr/.style={-{Latex[length=2mm]}, line width=0.7pt}
]

\node[font=\footnotesize\bfseries] (titleA) at (2.8, 5.2) {(a) Flat baseline / ClassicalMoE\_Shared};

\node[box, fill=red!25] (a1) at (0, 3.7) {Backbone\\(retrain)};
\node[box, fill=red!25] (a2) at (2.8, 3.7) {Existing\\heads (retrain)};
\node[box, fill=red!25] (a3) at (5.6, 3.7) {Gate / output\\dim (retrain)};
\node[box, fill=red!35] (a4) at (2.8, 2.1) {New crop\\class(es) added};

\draw[arr] (a1.south) -- ++(0,-0.3) -| (a4.north);
\draw[arr] (a2) -- (a4);
\draw[arr] (a3.south) -- ++(0,-0.3) -| (a4.north);

\node[align=center, below=0.35cm of a4, font=\scriptsize\itshape] (capA) {Entire system retrained\\($\sim$40 min/seed)};

\node[font=\footnotesize\bfseries] (titleB) at (11.1, 5.2) {(b) AdapterMoE (\texttt{add\_crop})};

\node[box, fill=green!20] (b1) at (8.3, 3.7) {Backbone\\(frozen, untouched)};
\node[box, fill=green!20] (b2) at (11.1, 3.7) {Existing Adapters\\(untouched)};
\node[box, fill=yellow!35] (b3) at (13.9, 3.7) {Router + DirectHead\\(light fine-tune)};
\node[box, fill=orange!35] (b4) at (11.1, 2.1) {\textbf{+1 new Adapter}\\(trained fresh)};
\node[box, fill=yellow!35] (b5) at (13.9, 2.1) {KNN bank\\(incremental update)};

\draw[arr] (b3.south) -- ++(0,-0.3) -| (b4.north);
\draw[arr] (b3) -- (b5);
\node[align=center, below=0.35cm of b4, font=\scriptsize\itshape] (capB) {Local operation\\($\sim$1.14 min/crop)};

\draw[dashed, gray] (6.9, 5.6) -- (6.9, 1.0);

\end{tikzpicture}}
\caption{Adding a new crop. (a) Flat baselines and the soft-routing MoE baseline must retrain the backbone, existing heads, and the output layer or gate end to end. (b) AdapterMoE's \texttt{add\_crop} interface freezes the backbone and existing Adapters, training only one new Adapter and lightly fine-tuning the Router and DirectHead.}
\label{fig:addcrop}
\end{figure*}

To support the realistic scenario of new crops arriving after deployment, we expose an \texttt{add\_crop} interface whose core principle is that \emph{existing Adapter weights never change}. Concretely, it runs four steps: (1) instantiate a new per-crop Adapter and train it only on the new crop's sub-dataset; (2) expand the Router's output layer from $K$ to $K{+}1$ classes and fine-tune it on the union of existing crops, the new crop, and Others; (3) similarly fine-tune DirectHead, whose flat output space grows accordingly; and (4) incrementally extend the KNN feature bank with the new crop's backbone features. Because the backbone stays frozen throughout, every step operates purely on the cached 1280-d feature space (\Cref{fig:addcrop}).

In our three-system comparison, \texttt{add\_crop} is unique to AdapterMoE: the flat baselines must expand their output layer and retrain end-to-end, and ClassicalMoE\_Shared's gate allocation is disrupted by a new crop and likewise requires full retraining. Hard routing turns ``add a crop'' into an additive operation rather than a redo -- we view this as the single most practically consequential difference between hard and soft routing exposed by our experiments (quantified in Section~\ref{sec:addcrop-results}).

\subsection{Five-System Fair Comparison Framework}
\label{sec:fair-comparison}

Our central claim is that two-stage hard routing dominates soft routing and monolithic flat classification on three axes -- training cost, new-crop extensibility, and rejection stability -- not on accuracy. Establishing this requires all systems to be evaluated under one task definition, one data configuration, and one evaluation protocol.

\textbf{Flat classification baselines.} We implement three variants trained directly on the flat 26-class configuration with no routing or decomposition: \emph{SimpleCNN}, a four-layer convolutional network with global average pooling ($\sim$0.10M parameters) included purely to establish a task-difficulty floor; \emph{ResNet-18}~\cite{he2016resnet} ($\sim$11.2M parameters) with a linear ($512{\to}26$) head; and \emph{EfficientNet-B0}~\cite{tan2019efficientnet} ($\sim$4.04M parameters) with a linear ($1280{\to}26$) head, both ImageNet-pretrained. All three train with Adam ($\mathrm{lr}=10^{-3}$) for 40 epochs with no early stopping, and reject Others via the same $\mathrm{MSP}<\theta=0.55$ rule as AdapterMoE.

\textbf{ClassicalMoE\_Shared (soft-routing MoE baseline).} This is the original dense-gating MoE of Jacobs et al.~\cite{jacobs1991moe} -- not Shazeer et al.'s sparse Top-$K$ variant~\cite{shazeer2017} and not Puigcerver et al.'s token-level Soft MoE~\cite{puigcerver2023softmoe} -- with five experts, one per target crop, and \emph{every} expert participating in every forward pass at a learned weight. A shared ImageNet-pretrained ResNet-18 backbone (trained end-to-end, unlike AdapterMoE's frozen backbone) feeds five independent linear Expert heads ($512{\to}26$ each) and a linear Gating head ($512{\to}5$); the final prediction is $y=\sum_i g_i\cdot\mathrm{Expert}_i(x)$ with $g=\mathrm{softmax}(\mathrm{Gating}(x))$. Total parameters ($\sim$11.25M) closely match AdapterMoE's 11.31M, supporting a like-for-like comparison at matched capacity. To counter expert collapse, we enable a load-balancing auxiliary loss (coefficient 0.01) following Shazeer et al.~\cite{shazeer2017}, $\mathcal{L}_{\mathrm{LB}}=\gamma^2(\text{importance}) + \gamma^2(\text{load})$, where importance is each expert's total gate weight within a batch, load is each expert's Top-1 selection count, and $\gamma$ is the coefficient of variation (std/mean); we write the full variant name as ClassicalMoE\_Shared\_LB but abbreviate to ClassicalMoE\_Shared throughout, and validate the necessity of this loss directly in Section~\ref{sec:ablation-lb}. Notably, ClassicalMoE\_Shared has no dedicated Others-rejection mechanism: Others is simply one of the 26 classes an Expert Head can output -- an inherent limitation of this soft-routing design and a key point of contrast with AdapterMoE's explicit two-stage rejection.

\textbf{AdapterMoE training pipeline.} The main system trains in five stages: Stage 0 backbone pre-calibration and freezing (3 epochs, $\sim$3.15 min/seed); Stage 1 RouterHead ($\sim$0.15 min); Stage 2 the five Adapters ($\sim$0.20 min combined, via feature caching); Stage 3 DirectHead ($\sim$0.04 min); and Stage 4 Temperature Scaling and OOD calibration for all seven branches plus Energy/KNN threshold fitting ($\sim$0.05 min). Table~\ref{tab:system-comparison} summarizes the four defining differences among the three system families.

\begin{table*}[t]
\centering
\small
\caption{Defining differences among the three system families compared in this work.}
\label{tab:system-comparison}
\begin{tabular}{@{}p{3.2cm}p{2.0cm}p{4.0cm}p{4.2cm}@{}}
\specialrule{1.1pt}{0pt}{0pt}
& \textbf{Flat baselines} & \textbf{ClassicalMoE\_Shared} & \textbf{AdapterMoE} \\
\specialrule{1.1pt}{0pt}{0pt}
Routing & None & Soft (learned gate) & Hard (deterministic Router) \\
\midrule
Others rejection & MSP post-hoc & No dedicated mechanism & Explicit Router-stage rejection \\
\midrule
Expert collapse risk & N/A & Present; needs LB loss & Precluded by design \\
\midrule
New-crop training cost & Full retrain & Full retrain & One Adapter + light fine-tune \\
\specialrule{1.1pt}{0pt}{0pt}
\end{tabular}
\end{table*}

\textbf{Unified task and evaluation split.} All five systems solve the identical task -- output one of 25 disease classes, or Others -- and are evaluated on the same flat 26-class test split ($N{=}1{,}999$; Others $=681$ samples, $\approx34\%$; the remaining 25 classes each have 50 samples except three Corn classes with slightly larger raw counts of 59, 77, and 82), fixed identically across all five seeds. Test-set metrics are computed once per run with no test-time hyperparameter tuning; all systems share identical augmentation, optimizer, batch size, and seed schedule, so the only varying factor is architecture itself.

\subsection{Training Configuration and Evaluation Metrics}
\label{sec:metrics}

All systems use Adam~\cite{kingma2015adam} ($\mathrm{lr}=10^{-3}$, weight decay $10^{-4}$) with Cosine Annealing~\cite{loshchilov2017sgdr} ($\eta_{\min}=10^{-5}$), batch size 32, and $224{\times}224$ inputs; early stopping is disabled everywhere in the main pipeline. Key threshold constants are shared across systems: $\theta_{\text{reject}}=\theta_{\text{high}}=0.55$, Energy/KNN thresholds set at a 0.95 true-positive-rate quantile on in-distribution validation data, and fusion weight $\alpha=0.6$. We do not employ AugMix~\cite{hendrycks2020augmix} or CutMix~\cite{yun2019cutmix} in the reported experiments; both are mature, well-validated augmentation techniques and we list their integration as future work rather than as part of the present system.

We report four primary classification metrics. Disease Accuracy restricts to the 25 named disease classes (excluding Others):
\begin{equation}
\dacc = \frac{1}{N_{\text{disease}}}\sum_{i\in\text{Disease}} \mathbf{1}[\hat{y}_i=y_i],
\label{eq:dacc}
\end{equation}
where $\mathbf{1}[\cdot]$ is the indicator function, equal to 1 if its argument holds and 0 otherwise. Others Accuracy measures rejection recall on the Others class:
\begin{equation}
\oacc = \frac{1}{N_{\text{others}}}\sum_{i\in\text{Others}} \mathbf{1}[\hat{y}_i=\text{Others}].
\label{eq:oacc}
\end{equation}
Overall Accuracy is computed across all 26 classes, and Macro-F1 -- our headline metric, robust to class imbalance -- averages per-class F1 unweighted over $K=26$ classes:
\begin{equation}
\macrof = \frac{1}{K}\sum_{c=1}^{K} \mathrm{F1}_c, \qquad \mathrm{F1}_c = \frac{2 P_c R_c}{P_c+R_c},
\label{eq:macrof1}
\end{equation}
where $P_c$ and $R_c$ are the precision and recall of class $c$ computed in the standard way (i.e., $P_c$ is the fraction of samples predicted as class $c$ that truly belong to class $c$, and $R_c$ is the fraction of true class-$c$ samples correctly predicted as class $c$). For calibration quality we report Expected Calibration Error (ECE, 15 bins)~\cite{naeini2015ece} and NLL (Section~\ref{sec:calibration}), and for the OOD gate, true/false positive rates and the Area Under the ROC Curve (AUROC).

\section{Experiments}
\label{sec:experiments}

\subsection{Setup}
\label{sec:setup}
All experiments run on a single machine (NVIDIA RTX 4070, 12GB; Intel Core i5-13500; 64GB DDR5 RAM; PyTorch 2.5.1+cu121; Python 3.10.19), with all randomness sources fixed and deterministic kernels enabled. Five full training-and-evaluation repetitions (seeds $\{42,123,456,789,2026\}$ -- chosen arbitrarily and fixed in advance, not individually meaningful) across the main pipeline, all baselines, and every ablation total approximately 820 GPU-minutes ($\approx$13.7 hours); every number reported below is an independent from-scratch run under this budget, with no cross-version weight reuse or extrapolation.

\subsection{Main System Comparison}
\label{sec:main-results}

\Cref{tab:main-results} reports the five systems' core performance, aggregated as mean $\pm$ std over five seeds on the shared $N{=}1{,}999$ flat 26-class test split.

\begin{table*}[t]
\centering
\footnotesize
\setlength{\tabcolsep}{4pt}
\caption{Main results: five systems, five-seed mean $\pm$ std on the shared test split. Bold marks AdapterMoE's two standout metrics (Train time, \oacc), discussed in the text; all other cells are reported as plain numbers to avoid implying an accuracy ranking that this paper does not claim (see Section~\ref{sec:main-results}).}
\label{tab:main-results}
\begin{tabular}{@{}lccccccc@{}}
\toprule
System & Macro-F1$\uparrow$ & Overall$\uparrow$ & \dacc$\uparrow$ & \oacc$\uparrow$ & Train (min)$\downarrow$ & Params (M)$\downarrow$ & Infer (ms)$\downarrow$ \\
\midrule
SimpleCNN & 74.50$\pm$1.01 & 74.30$\pm$1.14 & 77.85$\pm$1.21 & 67.43$\pm$2.98 & 40.44$\pm$3.22 & 0.10 & 0.75$\pm$0.10 \\
ResNet-18 & 97.60$\pm$0.46 & 98.13$\pm$0.39 & 97.60$\pm$0.38 & 99.15$\pm$0.59 & 39.89$\pm$2.12 & 11.19 & 2.56$\pm$0.10 \\
EfficientNet-B0 & 97.68$\pm$0.16 & 97.98$\pm$0.29 & 98.09$\pm$0.12 & 97.77$\pm$0.93 & 43.57$\pm$2.97 & 4.04 & 7.53$\pm$0.32 \\
ClassicalMoE\_Shared & 97.75$\pm$0.46 & 98.23$\pm$0.34 & 97.63$\pm$0.42 & 99.38$\pm$0.32 & 40.31$\pm$2.10 & 11.25 & 2.85$\pm$0.15 \\
AdapterMoE (ours) & 97.51$\pm$0.34 & 98.06$\pm$0.35 & 97.33$\pm$0.48 & \textbf{99.47}$\pm$0.17 & \textbf{3.59}$\pm$0.23 & 11.31 & 7.56$\pm$0.16 \\
\bottomrule
\end{tabular}

\vspace{2pt}
{\raggedright\footnotesize Note: AdapterMoE's \dacc\ (97.33) is numerically the lowest among the four strong systems, not the highest -- reported plainly rather than obscured. Params and Infer are shown for reference: SimpleCNN's low values reflect low capacity and should be read alongside its far lower accuracy, not as a favorable trade-off.\par}
\end{table*}

Three observations stand out. First, the four strong systems (ResNet-18, EfficientNet-B0, ClassicalMoE\_Shared, AdapterMoE) land within a 0.24-point Macro-F1 band (97.51--97.75); ClassicalMoE\_Shared's edge over AdapterMoE (0.24 points) is smaller than ClassicalMoE\_Shared's own cross-seed standard deviation (0.46), and AdapterMoE's gaps to ResNet-18 (0.09) and EfficientNet-B0 (0.08) are smaller still -- all four differences sit inside statistical noise. On this controlled benchmark, accuracy has saturated to the point of losing discriminative power, which is our explicit reason for not building this paper's argument around an accuracy claim. Second, AdapterMoE's training cost is a clear outlier in the good direction: $3.59\pm0.23$ minutes per seed against $40.31$ (ClassicalMoE\_Shared), $39.89$ (ResNet-18), and $43.57$ (EfficientNet-B0) -- respectively $8.9\%$, $9.0\%$, and $8.2\%$ of those costs, an $11$--$12\times$ efficiency gain, achieved because only 7.30M trainable parameters (RouterHead, five Adapters, DirectHead, and seven temperature scalars) are ever updated, and every image passes through the backbone exactly once across the entire training process (Section~\ref{sec:cost} discusses precisely what this efficiency gain should and should not be attributed to). Third, AdapterMoE's \oacc\ is simultaneously the highest of all five systems ($99.47\%$) and the most stable ($\mathrm{std}=0.17\%$, the lowest of the five) -- $3.5\times$ more stable than ResNet-18, $5.5\times$ more stable than EfficientNet-B0, and $1.9\times$ more stable than ClassicalMoE\_Shared.

\subsection{Training Cost}
\label{sec:cost}

\Cref{tab:cost} restates training cost relative to two natural reference points.

\begin{table*}[t]
\centering
\small
\caption{Training cost relative to ClassicalMoE\_Shared and to ResNet-18 (five-seed mean $\pm$ std; independent from-scratch measurements).}
\label{tab:cost}
\begin{tabular}{@{}lccc@{}}
\toprule
System & Train (min)$\downarrow$ & vs.\ ClassicalMoE\_Shared & vs.\ ResNet-18 \\
\midrule
SimpleCNN & 40.44$\pm$3.22 & 100.3\% & 101.4\% \\
ResNet-18 & 39.89$\pm$2.12 & 99.0\% & 100\% (ref.) \\
EfficientNet-B0 & 43.57$\pm$2.97 & 108.1\% & 109.2\% \\
ClassicalMoE\_Shared & 40.31$\pm$2.10 & 100\% (ref.) & 101.1\% \\
AdapterMoE (ours) & \textbf{3.59}$\pm$0.23 & \textbf{8.9\%} & \textbf{9.0\%} \\
\bottomrule
\end{tabular}
\end{table*}

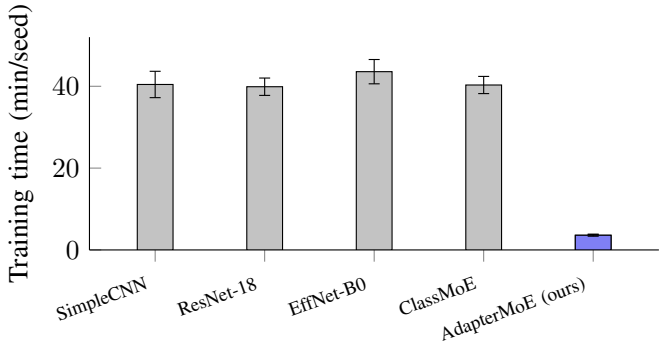
\begin{figure}[t]
\centering
\resizebox{\columnwidth}{!}{\begin{tikzpicture}
\begin{axis}[
    width=\columnwidth, height=4.3cm,
    ybar,
    bar width=13pt,
    ylabel={Training time (min/seed)},
    symbolic x coords={SimpleCNN,ResNet-18,EffNet-B0,ClassMoE,AdapterMoE},
    xtick=data,
    xticklabels={SimpleCNN,ResNet-18,EffNet-B0,ClassMoE,{AdapterMoE (ours)}},
    x tick label style={font=\scriptsize,rotate=20,anchor=north east,xshift=-2pt},
    ytick style={font=\scriptsize},
    nodes near coords style={font=\scriptsize},
    ymin=0, ymax=52,
    axis lines*=left,
    enlarge x limits=0.15,
    error bars/y dir=both, error bars/y explicit,
    every axis plot/.append style={fill opacity=0.85},
]
\addplot[fill=gray!55, bar shift=0pt, error bars/error bar style={black}] coordinates {
    (SimpleCNN,40.44) +- (0,3.22)
    (ResNet-18,39.89) +- (0,2.12)
    (EffNet-B0,43.57) +- (0,2.97)
    (ClassMoE,40.31) +- (0,2.10)
    (AdapterMoE,3.59) +- (0,0.23)
};
\addplot[fill=blue!55, bar shift=0pt, error bars/error bar style={black}] coordinates {
    (AdapterMoE,3.59) +- (0,0.23)
};
\end{axis}
\end{tikzpicture}}
\caption{Training cost by system (mean $\pm$ std, five seeds). AdapterMoE's advantage comes from frozen-backbone feature caching, not from hard routing per se (see caveat above).}
\label{fig:cost-chart}
\end{figure}

\textbf{A necessary caveat.} This cost gap primarily reflects a difference in \emph{training paradigm} -- AdapterMoE uses a frozen backbone with feature caching and updates only lightweight heads, while every other system fine-tunes its full network end to end -- and not, by itself, a difference between hard and soft \emph{routing}. The roughly $12\times$ figure should therefore be read as the advantage of ``frozen backbone plus feature caching'' over ``end-to-end fine-tuning,'' a training-paradigm axis that is largely orthogonal to the hard-vs-soft routing axis. Quantifying the impact of routing strategies on training costs separately would require a controlled comparison in which \emph{both} systems freeze their backbone -- an experiment we did not run and flag explicitly as future work (Section~\ref{sec:future}). \Cref{fig:cost-chart} visualizes this same comparison directly.

\subsection{Rejection Stability}
\label{sec:rejection-stability}

\Cref{tab:stability} restates \oacc\ stability directly, expressed as a multiple of AdapterMoE's own standard deviation.

\begin{table}[t]
\centering
\small
\caption{Cross-seed rejection stability: \oacc\ mean, standard deviation, and the ratio of each system's std to AdapterMoE's.}
\label{tab:stability}
\begin{tabular}{@{}lccc@{}}
\toprule
System & \oacc$\uparrow$ & \oacc\ std & std ratio \\
\midrule
SimpleCNN & 67.43\% & 2.98\% & 17.5$\times$ \\
ResNet-18 & 99.15\% & 0.59\% & 3.5$\times$ \\
EfficientNet-B0 & 97.77\% & 0.93\% & 5.5$\times$ \\
ClassicalMoE\_Shared & 99.38\% & 0.32\% & 1.9$\times$ \\
AdapterMoE (ours) & \textbf{99.47\%} & \textbf{0.17\%} & 1.0$\times$ (ref.) \\
\bottomrule
\end{tabular}
\end{table}

\begin{figure}[t]
\centering
\resizebox{\columnwidth}{!}{\begin{tikzpicture}
\begin{axis}[
    width=\columnwidth, height=4.3cm,
    ybar,
    bar width=13pt,
    ylabel={Others Accuracy (\%)},
    symbolic x coords={SimpleCNN,ResNet-18,EffNet-B0,ClassMoE,AdapterMoE},
    xtick=data,
    xticklabels={SimpleCNN,ResNet-18,EffNet-B0,ClassMoE,{AdapterMoE (ours)}},
    x tick label style={font=\scriptsize,rotate=20,anchor=east},
    tick label style={font=\scriptsize},
    label style={font=\scriptsize},
    ymin=55, ymax=105,
    axis lines*=left,
    enlarge x limits=0.15,
    error bars/y dir=both, error bars/y explicit,
]
\addplot[fill=gray!55, bar shift=0pt, error bars/error bar style={black}] coordinates {
    (SimpleCNN,67.43) +- (0,2.98)
    (ResNet-18,99.15) +- (0,0.59)
    (EffNet-B0,97.77) +- (0,0.93)
    (ClassMoE,99.38) +- (0,0.32)
    (AdapterMoE,99.47) +- (0,0.17)
};
\addplot[fill=blue!55, bar shift=0pt, error bars/error bar style={black}] coordinates {
    (AdapterMoE,99.47) +- (0,0.17)
};
\end{axis}
\end{tikzpicture}}
\caption{Others Accuracy by system, with error bars showing cross-seed standard deviation. AdapterMoE combines the highest mean with the tightest spread.}
\label{fig:stability-chart}
\end{figure}
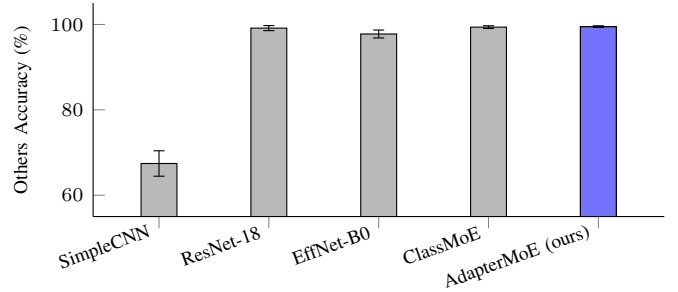

The practical stakes of cross-environment stability are concrete: differing field lighting, hardware-specific numerical error, and differing deployment-version floating-point behavior are all, statistically, akin to varying the random seed. A larger \oacc\ standard deviation means rejection behavior drifts when the deployment environment changes, undermining any commitment to a bounded false-positive rate; AdapterMoE's lowest-of-five standard deviation indicates its rejection boundary is the least sensitive to random initialization among the systems compared, which is what makes a deployment service-level commitment on rejection behavior plausible in the first place. \Cref{fig:stability-chart} plots this comparison, with error bars showing the cross-seed spread.

\subsection{Path Routing and Inference Latency}
\label{sec:pathrouting}

\Cref{tab:pathrouting} reports AdapterMoE's usage rate and conditional accuracy for each of its four decision paths.

\begin{table*}[t]
\centering
\small
\caption{AdapterMoE path usage and conditional accuracy (five-seed mean $\pm$ std).}
\label{tab:pathrouting}
\begin{tabular}{@{}lcc@{}}
\toprule
Path & Usage rate & Conditional accuracy \\
\midrule
Path 1 -- High (Adapter-led) & 65.73$\pm$0.16\% & 97.63$\pm$0.27\% \\
Path 2 -- Low (fusion) & 0.02$\pm$0.03\% & -- \\
Path 0 -- Reject (Others-direct) & 33.79$\pm$0.32\% & 99.59$\pm$0.28\% \\
Path 3 -- OOD-reject (dual-gate) & 0.46$\pm$0.20\% & -- \\
\bottomrule
\end{tabular}
\end{table*}

Two points follow. First, the Path-0 usage rate ($33.79\%$) closely tracks Others' true test-set share ($681/1{,}999\approx34.07\%$), indicating the Router discriminates Others almost perfectly on this benchmark. Second, Path 2 (the fusion path) triggers on only $0.02\%$ of samples -- on PlantVillage, the dual-gate-plus-DirectHead machinery is almost never exercised in the main pipeline. We read this plainly as DirectHead degenerating, on this particular benchmark, into an \emph{insurance} mechanism: normally inactive, but present so that the system still has a fallback decision-maker when the Router fails at a genuine boundary case; we expect its value to become visible specifically in deployments with more pronounced distribution shift, which we probe directly in Section~\ref{sec:plantdoc}.

On latency, AdapterMoE's per-image inference time is $7.56\pm0.16$ ms, of which the backbone forward pass accounts for $\approx7.11$ ms ($\approx94\%$), with RouterHead ($0.21$ ms) and the Adapter ($0.24$ ms) contributing the rest and DirectHead adding negligibly (a linear layer on an already-computed frozen feature). This is nearly identical to plain EfficientNet-B0's $7.53$ ms -- the shared-backbone design means adding five Adapters and a RouterHead costs only $0.03$ ms of additional latency, supporting AdapterMoE's viability for edge deployment.

\subsection{Incremental Learning with \texttt{add\_crop}}
\label{sec:addcrop-results}

We simulate the realistic scenario of crops arriving one at a time after deployment, adding Apple $\to$ Corn $\to$ Grape $\to$ Potato $\to$ Tomato in sequence to an initially empty AdapterMoE instance and tracking both the time each step takes and whether previously added crops' recognition degrades. \Cref{tab:addcrop} reports the five-seed average.

\begin{table*}[t]
\centering
\small
\caption{Five-step \texttt{add\_crop} expansion (five-seed mean $\pm$ std, except Step time).}
\label{tab:addcrop}
\begin{tabular}{@{}clccccc@{}}
\toprule
Step & Crop added & \dacc\ (\%) & \oacc\ (\%) & Overall (\%) & Macro-F1 (\%) & Step time (min) \\
\midrule
1 & Apple & 14.92$\pm$0.19 & 99.65$\pm$0.17 & 43.78$\pm$0.14 & 17.25$\pm$0.13 & 4.30 \\
2 & +Corn & 34.61$\pm$0.33 & 99.77$\pm$0.08 & 56.81$\pm$0.21 & 32.45$\pm$0.22 & 1.11 \\
3 & +Grape & 49.71$\pm$0.19 & 99.74$\pm$0.19 & 66.75$\pm$0.18 & 47.99$\pm$0.17 & 1.10 \\
4 & +Potato & 60.97$\pm$0.32 & 99.62$\pm$0.30 & 74.14$\pm$0.22 & 59.56$\pm$0.25 & 1.08 \\
5 & +Tomato & 97.50$\pm$0.39 & 99.35$\pm$0.17 & 98.13$\pm$0.24 & 97.63$\pm$0.33 & 1.29 \\
\bottomrule
\end{tabular}
\end{table*}

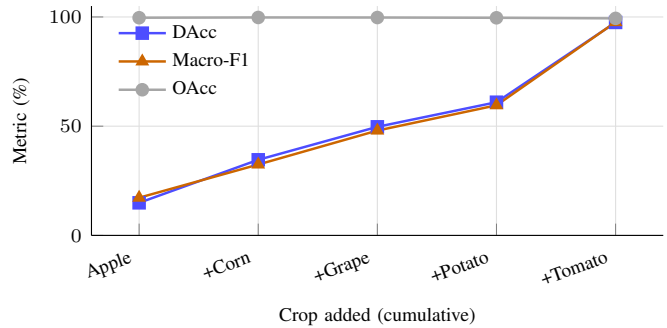
\begin{figure}[t]
\centering
\resizebox{\columnwidth}{!}{\begin{tikzpicture}
\begin{axis}[
    width=\columnwidth, height=4.5cm,
    xlabel={Crop added (cumulative)},
    ylabel={Metric (\%)},
    xtick={1,2,3,4,5},
    xticklabels={Apple,+Corn,+Grape,+Potato,+Tomato},
    x tick label style={font=\scriptsize,rotate=20,anchor=north east,xshift=-2pt},
    tick label style={font=\scriptsize},
    label style={font=\scriptsize},
    ymin=0, ymax=105,
    axis lines*=left,
    grid=major, grid style={gray!25},
    legend style={font=\scriptsize,at={(0.03,0.97)},anchor=north west,draw=none,fill=none},
    legend cell align={left},
]
\addplot[color=blue!70, mark=square*, line width=0.9pt] coordinates {
    (1,14.92)(2,34.61)(3,49.71)(4,60.97)(5,97.50)
};
\addlegendentry{DAcc}
\addplot[color=orange!80!black, mark=triangle*, line width=0.9pt] coordinates {
    (1,17.25)(2,32.45)(3,47.99)(4,59.56)(5,97.63)
};
\addlegendentry{Macro-F1}
\addplot[color=gray!70, mark=*, line width=0.9pt] coordinates {
    (1,99.65)(2,99.77)(3,99.74)(4,99.62)(5,99.35)
};
\addlegendentry{OAcc}
\end{axis}
\end{tikzpicture}}
\caption{Metric trajectory during \texttt{add\_crop} expansion. \dacc\ and Macro-F1 rise as crops are added (by design, since not-yet-added crops are correctly routed to Others); \oacc\ stays flat, showing existing crops are undisturbed.}
\label{fig:addcrop-curve}
\end{figure}

Two conclusions follow, plotted directly in \Cref{fig:addcrop-curve}. First, expanding by one crop costs $1.14$ minutes on average (Steps 2--5, range $1.08$--$1.29$), $\approx35\times$ faster than ClassicalMoE\_Shared's $40.31$-minute from-scratch retrain and $\approx35\times$ faster than ResNet-18's $39.89$ minutes; Step 1 (Apple) takes longer ($4.30$ min) because it must additionally build the KNN bank and initialize the Router's training data from scratch. Second, \oacc\ stays in a tight $99.35$--$99.77\%$ band throughout expansion -- adding crops barely perturbs existing crops' rejection behavior. The low \dacc\ in Steps 1--4 is by design, not a defect: crops not yet added are, correctly, treated by the Router as non-target and routed to rejection, and this rising curve is itself evidence that \texttt{add\_crop} behaves as intended; by Step 5, with all five crops present, \dacc\ reaches $97.50\%$, matching the main pipeline's $97.33\%$ within cross-seed standard deviation. A seed-level check (comparing Step-5 \dacc\ against the from-scratch main pipeline's \dacc\ for each of the five seeds) shows 4 of 5 seeds at or above the main pipeline's accuracy (mean $+0.16$ points), with seed 789 the sole exception ($-0.68$ points) -- a discrepancy we trace in Section~\ref{sec:crossseed} to that seed's OOD gate happening to perform unusually well in the from-scratch run specifically, an advantage tied to a particular backbone-and-KNN-bank state that does not carry over once the KNN bank is rebuilt incrementally.

\subsection{Ablation Studies}
\label{sec:ablation}

We report eight ablations, each run under the full five-seed protocol. The first four validate core design choices in the main system; the remaining four are controlled comparisons that isolate specific claims (Adapter irreplaceability, the necessity of the load-balancing loss, and threshold stability).

\subsubsection{Tomato Adapter capacity}
\label{sec:ablation-tomato}
Tomato is the heaviest sub-task (10 classes) and the largest Adapter, making it the most overfitting-prone branch. We compare the standard architecture (Std\_r8) against three enhanced variants (Enh\_r4, Enh\_r8, Enh\_r16, where $r$ is the SE reduction ratio), reported in \Cref{tab:tomato}.

\begin{table}[t]
\centering
\small
\caption{Tomato Adapter variants (five-seed mean $\pm$ std). Flat F1 is computed after mapping to the 26-class space. Enh\_r4 (*) is the main-pipeline choice.}
\label{tab:tomato}
\begin{tabular}{@{}lccc@{}}
\toprule
Variant & Flat F1 & Params (M) & $T$ \\
\midrule
Enh\_r16 & 0.9598$\pm$0.0075 & 2.05 & 0.803$\pm$0.064 \\
Enh\_r8 & 0.9593$\pm$0.0061 & 2.26 & 0.826$\pm$0.086 \\
Enh\_r4* & 0.9577$\pm$0.0045 & 2.67 & 0.805$\pm$0.103 \\
Std\_r8 & 0.9578$\pm$0.0064 & 1.07 & 2.184$\pm$0.650 \\
\bottomrule
\end{tabular}
\end{table}

Three findings. First, the enhanced variants lead Std\_r8 by only $\approx0.19$ points on L2 Tomato F1, a gap smaller than either variant's own cross-seed standard deviation -- accuracy alone does not justify the added capacity. Second, the enhanced variants' real advantage is calibration quality: Std\_r8's Tomato temperature is $2.184\pm0.650$ (severely overconfident and unstable across seeds), while all three enhanced variants sit near $T\approx0.80$ with roughly one-sixth the cross-seed standard deviation, plausibly attributable to the enhanced variant's label smoothing. Third, on parameter cost, Enh\_r4 (our main-pipeline choice) costs 1.60M more than Std\_r8; since the three enhanced variants are statistically indistinguishable from each other on accuracy, Enh\_r16 is a reasonable lower-parameter alternative, and Std\_r8 itself (using 40\% of Enh\_r4's parameters) remains defensible if inference latency is the deployment priority, since its accuracy gap is within noise.

\subsubsection{Dual-gate parameter scan: why we did not run a full grid}
\label{sec:ablation-gate}
We originally planned a three-dimensional grid scan over $(\theta_{\text{high}}, \alpha, \text{DirectHead on/off})$. An early pilot scan showed that, under the main PlantVillage configuration, the Low-Path usage rate stayed between $0.0\%$ and $0.4\%$ across every combination tested, meaning $\alpha$ and the DirectHead toggle had a measurable effect on essentially no samples -- all 24 grid combinations produced identical \dacc/\oacc/Overall/Macro-F1 to the precision we measured. We therefore stopped the full grid rather than spend compute confirming a null result already visible in the pilot, and instead report the main pipeline's directly measured Low-Path rate of $0.02\%$ (Section~\ref{sec:pathrouting}), consistent with the pilot. This null result does not mean the two-zone design is useless -- it means it is an \emph{insurance} mechanism whose value is invisible on a saturated, in-distribution benchmark and is expected to surface specifically when MSP is more broadly depressed (e.g., OOD-heavy or low-data deployments). We retain $\theta_{\text{high}}=0.55$, $\alpha=0.6$, and DirectHead enabled in the main pipeline because no better alternative was identified and because DirectHead remains useful as a fallback for \texttt{add\_crop} and for distribution-shifted deployment (Section~\ref{sec:plantdoc}).

\subsubsection{Temperature Scaling calibration}
\label{sec:ablation-calibration}
\Cref{tab:calib-T} reports ECE before/after calibration and the learned temperatures for all seven branches -- included specifically to make the calibration mechanism auditable rather than reporting only a summary improvement.

\begin{table*}[t]
\centering
\small
\caption{Calibration results per branch (five-seed mean $\pm$ std). $T$ is post-clamp (bounds $[0.3,10.0]$); \# low is how many of the 5 seeds hit the lower clamp bound.}
\label{tab:calib-T}
\begin{tabular}{@{}lccccc@{}}
\toprule
Branch & ECE before (\%) & ECE after (\%) & Reduction & $T$ (mean $\pm$ std) & \# low (of 5) \\
\midrule
Router & 4.88$\pm$0.16 & 0.55$\pm$0.22 & 88.9\% & 0.51$\pm$0.07 & 0/5 \\
Adapter\_Apple & 2.66$\pm$1.64 & 0.00 & 99.9\% & 0.33$\pm$0.07 & 4/5 \\
Adapter\_Corn & 5.12$\pm$1.36 & 4.31$\pm$1.22 & 15.8\% & 1.70$\pm$1.19 & 0/5 \\
Adapter\_Grape & 2.28$\pm$0.84 & 0.00 & 100.0\% & 0.31$\pm$0.01 & 2/5 \\
Adapter\_Potato & 7.77$\pm$1.31 & 0.00 & 100.0\% & 0.30$\pm$0.00 & 5/5 \\
Adapter\_Tomato & 4.47$\pm$1.41 & 2.52$\pm$0.69 & 40.2\% & 0.82$\pm$0.11 & 0/5 \\
DirectHead & 1.47$\pm$0.66 & 1.01$\pm$0.47 & 29.3\% & 0.80$\pm$0.11 & 0/5 \\
\bottomrule
\end{tabular}
\end{table*}

Calibration succeeds for six of seven branches (ECE reductions of 29--100\%), but Adapter\_Corn is a consistent failure across all five seeds: its temperature ranges from $0.65$ to $3.14$ (a $5\times$ spread) and post-calibration ECE remains at $4.31\%$. We attribute this to Corn's four classes containing at least one visually near-identical disease pair (\emph{Cercospora} leaf spot and Northern Leaf Blight), whose probability sharpness is plausibly not correctable by a single scalar $T$. We disclose this as a known limitation and list a targeted Corn Adapter redesign -- e.g., per-class temperature vectors or class-balanced sampling -- as future work (Section~\ref{sec:future}).

\subsubsection{OOD gate configuration}
\label{sec:ablation-ood}
\label{sec:ablation-ood-full}
We compare five OOD-gate configurations -- disabled, Energy-only, KNN-only, Energy \texttt{AND} KNN (main pipeline), and Energy \texttt{OR} KNN -- in \Cref{tab:oodgate}.

\begin{table*}[t]
\centering
\small
\caption{OOD gate configuration comparison (five-seed mean $\pm$ std).}
\label{tab:oodgate}
\begin{tabular}{@{}lcccc@{}}
\toprule
Variant & Macro-F1 & \dacc & \oacc & OOD-reject rate \\
\midrule
D1: disabled & 97.44$\pm$0.12 & 97.59$\pm$0.33 & 98.77$\pm$0.88 & 0.00\% \\
D2: Energy only & 96.33$\pm$0.43 & 94.89$\pm$0.75 & 99.91$\pm$0.08 & 2.38\% \\
D3: KNN only & 97.32$\pm$0.40 & 96.86$\pm$0.51 & 99.47$\pm$0.17 & 0.91\% \\
D4: Energy \texttt{AND} KNN (main) & 97.51$\pm$0.34 & 97.33$\pm$0.48 & 99.47$\pm$0.17 & 0.46\% \\
D5: Energy \texttt{OR} KNN & 96.13$\pm$0.39 & 94.42$\pm$0.72 & 99.91$\pm$0.08 & 2.83\% \\
\bottomrule
\end{tabular}
\end{table*}

D4's Macro-F1 gain over D1 (disabled) is only $+0.07$ points, within noise, and 4 of the 5 seeds show D4 no better than D1; the average advantage is driven almost entirely by a single seed (789, $+0.80$ points). Two conclusions follow. First, \texttt{AND} is the right combination logic: \texttt{OR} (D5) reaches $99.91\%$ \oacc\ but at the cost of over-rejection, dragging \dacc\ down to $94.42\%$ ($-2.91$ points versus D4); neither single signal alone matches \texttt{AND}'s robustness. Second, and more importantly, this result exposes the OOD gate's \emph{conditional} value: because its average benefit traces to one seed-789 outlier -- the same seed whose \texttt{add\_crop} run reverses direction (Section~\ref{sec:addcrop-results}) -- we do not claim the OOD gate is unconditionally better than disabling it under in-distribution-only evaluation; we expect its value to appear reliably only on genuinely out-of-distribution inputs, which we test directly in Section~\ref{sec:plantdoc}.

\subsubsection{Backbone pre-calibration epoch sweep}
\label{sec:ablation-epoch}
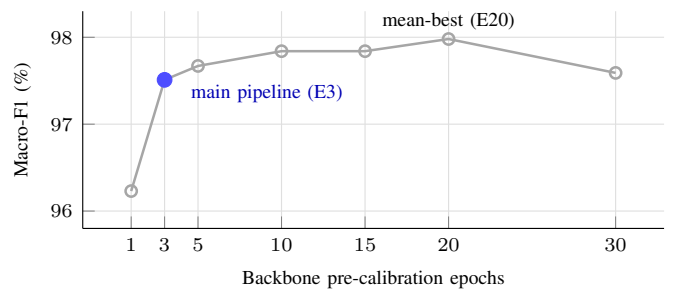
\begin{figure}[t]
\centering
\resizebox{\columnwidth}{!}{\begin{tikzpicture}
\begin{axis}[
    width=\columnwidth, height=4.3cm,
    xlabel={Backbone pre-calibration epochs},
    ylabel={Macro-F1 (\%)},
    xtick={1,3,5,10,15,20,30},
    ymin=95.8, ymax=98.3,
    label style={font=\scriptsize},
    tick label style={font=\scriptsize},
    axis lines*=left,
    grid=major, grid style={gray!25},
]
\addplot[color=gray!70, mark=o, mark options={fill=gray!70}, line width=0.9pt] coordinates {
    (1,96.23) (3,97.51) (5,97.67) (10,97.84) (15,97.84) (20,97.98) (30,97.59)
};
\addplot[color=blue!70, mark=*, mark options={fill=blue!70,scale=1.3}, only marks] coordinates {
    (3,97.51)
};
\node[font=\scriptsize,blue!70!black,anchor=west] at (axis cs:4,97.35) {main pipeline (E3)};
\node[font=\scriptsize,anchor=south] at (axis cs:20,97.98) {mean-best (E20)};
\end{axis}
\end{tikzpicture}}
\caption{Macro-F1 vs.\ backbone pre-calibration epochs (five-seed mean). The main pipeline uses E3 for speed; accuracy keeps rising through E20 before overfitting sets in at E30.}
\label{fig:epoch-sweep}
\end{figure}

The main pipeline uses 3 epochs of backbone pre-calibration before freezing, chosen for speed; \Cref{fig:epoch-sweep} sweeps 7 values to check whether this is actually a good trade-off point.

The mean-optimal point is 20 epochs ($+0.48$ Macro-F1 points over E3, at a cost of $+21$ min/seed); 30 epochs shows overfitting-driven regression. This is the clearest demonstration that AdapterMoE is not a fixed ``fast but capped'' point but a slidable cost--accuracy curve: a user willing to spend $+21$ min/seed can reach $97.98\%$ Macro-F1, exceeding ClassicalMoE\_Shared's $97.75\%$ main result. We note, however, that seed-level preferences diverge (2 of 5 seeds favor E15, 2 favor E20, 1 favors E30), so E20's mean-optimal status is an averaging effect, not a majority-vote result, and any future adoption of E20 as a new default would need re-validation on more seeds before being described as an unambiguous improvement.

\subsubsection{Control baseline: Adapter irreplaceability}
\label{sec:ablation-removal}
The core methodological assumption behind per-crop Adapters is that each one learns crop-specific fine-grained discrimination that cannot be substituted by the Router or by another Adapter. We test this by removing one Adapter at a time and re-routing its crop's samples through three fallback rules: \emph{DirectHead Fallback} (send the removed crop's samples to the flat 26-way DirectHead), \emph{Router\_top1 Fallback} (predict that crop's single most frequent training disease as a constant guess), and \emph{Router\_direct Fallback} (route through the Router's crop decision, then to DirectHead). \Cref{tab:removal} reports the resulting drop in \dacc\ for the affected crop.

\begin{table*}[t]
\centering
\small
\caption{\dacc\ drop (percentage points) after removing one Adapter, by fallback rule.}
\label{tab:removal}
\begin{tabular}{@{}lcccc@{}}
\toprule
Removed crop & \# classes & DirectHead FB & Router\_top1 FB & Router\_direct FB \\
\midrule
Apple & 4 & $-$14.98 & $-$11.31 & $-$15.07 \\
Corn & 4 & $-$19.74 & $-$13.54 & $-$19.74 \\
Grape & 4 & $-$14.85 & $-$10.96 & $-$14.83 \\
Potato & 3 & $-$11.21 & $-$7.56 & $-$11.31 \\
Tomato & 10 & $-$36.42 & $-$32.75 & $-$36.56 \\
\bottomrule
\end{tabular}
\end{table*}

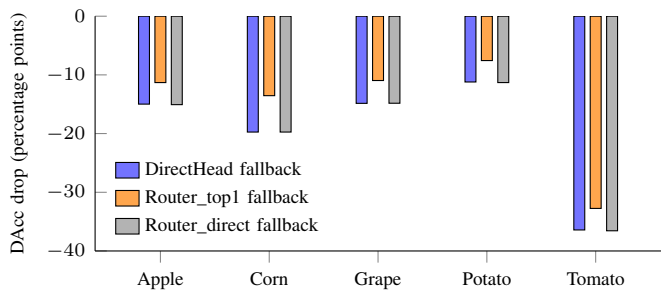
\begin{figure}[t]
\centering
\resizebox{\columnwidth}{!}{\begin{tikzpicture}
\begin{axis}[
    width=\columnwidth, height=4.6cm,
    ybar,
    bar width=4pt,
    ylabel={DAcc drop (percentage points)},
    symbolic x coords={Apple,Corn,Grape,Potato,Tomato},
    xtick=data,
    x tick label style={font=\scriptsize},
    tick label style={font=\scriptsize},
    label style={font=\scriptsize},
    ymin=-40, ymax=0,
    axis lines*=left,
    enlarge x limits=0.15,
    legend style={font=\scriptsize,at={(0.02,0.02)},anchor=south west,draw=none,fill=none},
    legend cell align={left},
    legend columns=1,
    legend image code/.code={\draw[#1] (0cm,-0.1cm) rectangle (0.3cm,0.1cm);},
]
\addplot[fill=blue!55] coordinates {
    (Apple,-14.98)(Corn,-19.74)(Grape,-14.85)(Potato,-11.21)(Tomato,-36.42)
};
\addlegendentry{DirectHead fallback}
\addplot[fill=orange!70] coordinates {
    (Apple,-11.31)(Corn,-13.54)(Grape,-10.96)(Potato,-7.56)(Tomato,-32.75)
};
\addlegendentry{Router\_top1 fallback}
\addplot[fill=gray!60] coordinates {
    (Apple,-15.07)(Corn,-19.74)(Grape,-14.83)(Potato,-11.31)(Tomato,-36.56)
};
\addlegendentry{Router\_direct fallback}
\end{axis}
\end{tikzpicture}}
\caption{DAcc drop after removing one Adapter, by fallback rule (\Cref{tab:removal}). Every fallback, for every crop, loses double digits -- no substitute comes close to the removed Adapter's function.}
\label{fig:removal-chart}
\end{figure}

\Cref{fig:removal-chart} plots this same result: every fallback, for every crop, produces a double-digit accuracy drop; Tomato's removal is the most damaging ($-36.42$ points), consistent with its ten-class burden. No fallback rule comes close to recovering the removed Adapter's function -- neither routing to the flat 26-way DirectHead nor falling back on the Router's crop-level decision substitutes for per-crop fine-grained discrimination. This directly supports the two-stage design (crop routing and disease classification are genuinely separate competencies) and, by extension, \texttt{add\_crop}'s premise that a new crop requires its own new Adapter rather than a re-purposed existing module.

\vspace{-10pt}
\subsubsection{Control baseline: load-balancing loss in ClassicalMoE\_Shared}
\label{sec:ablation-lb}
ClassicalMoE\_Shared enables a load-balancing (LB) auxiliary loss by default (coefficient 0.01). We test whether this is actually load-bearing by comparing it against an otherwise-identical variant with LB disabled (\Cref{tab:lbloss}); MaxGate is the largest of the five experts' gate probabilities (near 1 indicates single-expert dominance, i.e., collapse), and Gate-$\gamma$ is the coefficient of variation of the gate distribution.

\begin{table*}[t]
\centering
\small
\caption{ClassicalMoE\_Shared with and without the load-balancing loss (five-seed mean $\pm$ std).}
\label{tab:lbloss}
\begin{tabular}{@{}lcccc@{}}
\toprule
Variant & \dacc\ (\%) & Macro-F1 (\%) & MaxGate & Gate-$\gamma$ \\
\midrule
With LB (= ClassicalMoE\_Shared) & 96.36$\pm$2.27 & 93.04$\pm$2.04 & 0.350$\pm$0.116 & 0.524$\pm$0.286 \\
No LB & 97.69$\pm$0.16 & 94.07$\pm$0.16 & 0.807$\pm$0.160 & 1.585$\pm$0.308 \\
\bottomrule
\end{tabular}
\end{table*}

Disabling LB drives MaxGate from $0.35$ to $0.81$ (as high as $0.99$ for some seeds) -- soft-routing MoE reliably collapses onto a dominant expert without this constraint. Enabling LB suppresses MaxGate back to $0.35$, but at a cost of $1.03$ Macro-F1 points, revealing a genuine structural trade-off between enforced load balance and accuracy that a soft-routing design cannot escape: with LB, collapse is avoided but accuracy is taxed; without LB, accuracy improves but collapse returns. This validates our choice to enable LB in ClassicalMoE\_Shared (making it the honest strongest version of the soft-routing baseline), and more importantly demonstrates \emph{why} we adopt hard routing instead: AdapterMoE avoids collapse through an a-priori semantic decomposition (crop $\to$ disease) built into the architecture, not through an auxiliary loss applied after the fact to patch a design that would otherwise collapse.

\subsubsection{Control baseline: MSP threshold stability}
\label{sec:ablation-theta}
The main pipeline fixes the Others-rejection threshold at $\theta_{\text{reject}}=0.55$. A practical deployment risk is that this value might need delicate tuning and might not tolerate operator drift (e.g., a technician nudging it to 0.40 or 0.80). We swept $\theta\in[0.40,0.80]$ and computed \dacc's per-seed range (max $-$ min) for EfficientNet-B0 versus AdapterMoE: averaged across seeds, the flat baseline's \dacc\ swings by $8.99$ points as $\theta$ varies (as much as $15.02$ points for seed 2026), while AdapterMoE swings by only $0.47$ points -- roughly a $19\times$ difference in threshold sensitivity. This is a second, independent line of evidence for rejection stability, complementing Section~\ref{sec:rejection-stability}'s same-$\theta$-across-seeds result with a same-seed-across-$\theta$ result: together they show AdapterMoE's rejection behavior is robust both to random initialization and to operator-level threshold drift, both of which are realistic sources of variation across field deployments with different lighting, hardware, and personnel.

\subsection{Cross-Seed Variance}
\label{sec:crossseed}
\Cref{tab:crossseed} reports AdapterMoE's per-seed metrics to show the individual seed values underlying the mean $\pm$ std summaries used elsewhere. Seed 789 is the clear outlier: it reaches AdapterMoE's best Macro-F1 ($98.07\%$, $+1.65\sigma$) and is also the seed whose OOD gate most clearly helps (Section~\ref{sec:ablation-ood}); the same seed, however, reverses direction in the \texttt{add\_crop} experiment ($-0.68$ points versus its own from-scratch run, Section~\ref{sec:addcrop-results}). Taken together, this cross-validates our reading that seed 789's main-pipeline advantage stems from an incidental pairing of a particular backbone state with a particular KNN bank, not from a stable property of the architecture. Seed 42 is AdapterMoE's weakest seed ($97.18\%$, $-0.97\sigma$); we note this seed happens to favor ResNet-18-backboned systems (the flat ResNet-18 baseline and ClassicalMoE\_Shared), putting AdapterMoE's EfficientNet-B0-backboned pipeline at a relative disadvantage specifically on this seed. For comparison, ClassicalMoE\_Shared's own Macro-F1 standard deviation ($0.46$) is $1.35\times$ AdapterMoE's ($0.34$), driven largely by a seed-123 outlier ($97.03\%$, $-1.56\sigma$) whose training curve shows an early sharp dip -- consistent with soft-routing's gate allocation being more sensitive to random seed than hard routing's deterministic assignment -- and this is, incidentally, the one seed on which AdapterMoE actually beats ClassicalMoE\_Shared outright ($97.47\%$ vs.\ $97.03\%$).

\begin{table}[t]
\centering
\small
\caption{AdapterMoE per-seed results underlying the five-seed summaries reported elsewhere.}
\label{tab:crossseed}
\begin{tabular}{@{}lcccc@{}}
\toprule
Seed & Macro-F1 (\%) & \dacc\ (\%) & Overall (\%) & \oacc\ (\%) \\
\midrule
42 & 97.18 & 96.89 & 97.70 & 99.27 \\
123 & 97.47 & 97.19 & 97.95 & 99.41 \\
456 & 97.29 & 97.04 & 97.85 & 99.41 \\
789 & 98.07 & 98.10 & 98.60 & 99.56 \\
2026 & 97.54 & 97.42 & 98.20 & 99.71 \\
\midrule
mean & 97.51 & 97.33 & 98.06 & 99.47 \\
std & 0.34 & 0.48 & 0.35 & 0.17 \\
\bottomrule
\end{tabular}
\end{table}

\subsection{Qualitative Error Analysis}
\label{sec:qualitative}
On seed 42's test set, ClassicalMoE\_Shared and AdapterMoE agree, and are correct, on over 96\% of samples, indicating the two very different routing strategies converge to highly similar decisions on this controlled benchmark; successful cases cluster around clear lesion patterns with Router MSP near 1.00. Among the samples both systems misclassify, we identify four recurring failure patterns: (i) \emph{within-crop confusion at high confidence} -- e.g., Corn's Common Rust mistaken for \emph{Cercospora} leaf spot, or Tomato's Early Blight for Target Spot, both with $\mathrm{MSP}\approx1.00$, indicating the model can be confidently wrong on genuinely ambiguous within-crop pairs; (ii) \emph{OOD false rejection} -- in-distribution samples (e.g., a healthy Apple leaf at $\mathrm{MSP}=0.95$) incorrectly flagged as Others by the dual gate, reflecting conservative over-rejection; (iii) \emph{Others false acceptance} -- a true Others sample passing through at moderate confidence ($\mathrm{MSP}=0.63$) and being misclassified as a target-crop disease, reflecting a gap in the opposite direction; and (iv) \emph{Router cross-crop misrouting} -- a Tomato sample routed to the Apple Adapter at the Router stage itself, which then confidently (but wrongly) classifies it as an Apple disease. Notably, failure mode (i) occurs \emph{after} correct crop routing, meaning further improvement here requires strengthening within-crop fine-grained discrimination (e.g., the Tomato Target-Spot-specific direction in Section~\ref{sec:future}), not adjusting the routing mechanism itself. The recurrence of high-MSP errors across these cases is consistent with the accuracy-saturation picture of Section~\ref{sec:main-results}: even where the model is wrong, its confidence often remains high.

\subsection{External Zero-Shot Evaluation on PlantDoc}
\label{sec:plantdoc}

Sections~\ref{sec:main-results}--\ref{sec:qualitative} evaluate entirely within PlantVillage, whose Others class -- built from other PlantVillage crops such as pepper, strawberry, and cherry -- still shares PlantVillage's laboratory-condition visual statistics with the five target crops. A stricter test of the dual-gate mechanism requires images the model never saw during training and that were captured under genuinely different conditions. We therefore evaluate AdapterMoE, zero-shot and with no fine-tuning or domain adaptation, on PlantDoc~\cite{singh2020plantdoc}, a dataset of web- and field-sourced plant photographs.

\subsubsection{Motivation and design}
PlantDoc's background, lighting, leaf framing, and resolution differ sharply from PlantVillage's studio conditions, making it a direct probe of covariate-shift tolerance. Of PlantDoc's 28 classes, 19 map onto our Flat-26 label space (the \emph{in-mapping} subset) and 9 belong to crops outside our five targets (Bell pepper, Blueberry, Cherry, Peach, Raspberry, Soyabean, Squash, Strawberry, etc.), which we treat as genuine OOD samples. Because our training data contains zero PlantDoc images and we perform no PlantDoc-specific tuning, this is a strict zero-shot external evaluation.

\subsubsection{Dataset and class mapping}
Merging PlantDoc's train and test splits gives 2{,}922 images; after class mapping, the in-mapping subset totals 2{,}101 images (across the 19 non-Others classes) and the OOD subset totals 821 images. Six of our Flat-26 classes have no corresponding PlantDoc samples (e.g., Apple\_Black\_rot, Corn\_healthy) and are excluded from this evaluation; class correspondence was established via keyword matching on crop name and 1--2 distinguishing disease terms, then manually verified.

We report two independent metrics: the \emph{OOD-rejection rate} on the 821 OOD samples (fraction correctly output as Others, via either Path 0 or Path 3), and \emph{in-mapping cross-domain accuracy} on the 2{,}101 in-mapping samples (fraction correctly assigned to their true Flat-26 class; a false rejection here counts as an error).

\subsubsection{Result 1: dual-gate OOD rejection generalizes}
\Cref{tab:plantdoc-ood} reports the aggregate and per-class OOD-rejection rate.

\begin{table}[t]
\centering
\small
\caption{PlantDoc OOD-rejection rate, aggregate and by rejection path (five-seed mean $\pm$ std).}
\label{tab:plantdoc-ood}
\begin{tabular}{@{}lc@{}}
\toprule
Metric & Mean $\pm$ std \\
\midrule
Overall OOD-rejection rate (9 classes) & 95.69\% $\pm$ 2.78\% \\
\quad via Path 0 (Router $\to$ Others) & 77.59\% $\pm$ 8.30\% \\
\quad via Path 3 (Energy+KNN gate) & 18.10\% $\pm$ 6.95\% \\
\bottomrule
\end{tabular}
\end{table}

Per-class rejection rates range narrowly from $93.01\%$ (Bell pepper leaf spot) to $99.29\%$ (Peach leaf), with all 9 OOD classes above $93\%$ and a cross-seed standard deviation of only $2.78\%$ overall -- no OOD class fails systematically. Most rejection is handled by the Router itself ($77.59\%$); the dual OOD gate catches an additional $18.10\%$ of cases the Router routes to a target crop, the two mechanisms acting complementarily. This is, to our knowledge, the first evaluation of this dual-gate design against a genuinely external, field-condition OOD source rather than an in-distribution proxy, and it substantially strengthens the case that the gate generalizes beyond the benchmark it was tuned on.

\subsubsection{Result 2: in-mapping cross-domain classification degrades sharply}
\Cref{tab:plantdoc-inmap} reports in-mapping performance in aggregate; per-class accuracy ranges from Corn\_Gray\_leaf\_spot's outlying $61.49\pm15.56\%$ down to several classes near $0$--$5\%$ (e.g., Tomato\_leaf (healthy) at $0.38\pm0.86\%$), with Corn\_Gray\_leaf\_spot's relatively strong showing plausibly explained by its field images happening to share PlantVillage's characteristic large-leaf, texture-dominant framing more than most other classes do.

\begin{table}[t]
\centering
\small
\caption{PlantDoc in-mapping cross-domain classification (five-seed mean $\pm$ std).}
\label{tab:plantdoc-inmap}
\begin{tabular}{@{}lc@{}}
\toprule
Metric & Mean $\pm$ std \\
\midrule
In-mapping cross-domain accuracy & 11.01\% $\pm$ 1.13\% \\
Fraction falsely rejected & 68.60\% $\pm$ 5.86\% \\
\quad via Path 0 (Router $\to$ Others) & 41.97\% $\pm$ 6.10\% \\
\quad via Path 3 (OOD gate) & 26.62\% $\pm$ 5.00\% \\
Fraction reaching Path 1 (classified) & 31.09\% $\pm$ 5.83\% \\
\bottomrule
\end{tabular}
\end{table}

In-mapping accuracy of $11.01\%$ sits roughly 87 points below PlantVillage's own $97.70\%$, with $68.60\%$ of in-mapping samples falsely rejected rather than misclassified into a wrong disease.

\subsubsection{Failure mode: covariate shift and conservative rejection}
\label{sec:covariate-shift}
This degradation is consistent with a covariate shift between PlantVillage (clean or single-color background, leaf as the dominant subject, controlled uniform lighting, close-range single-leaf framing, standardized resolution) and PlantDoc (cluttered field background, leaf sometimes a small fraction of the frame, variable natural lighting with shadow, medium-to-long range and multi-leaf framing, inconsistent resolution). This gap is well documented in prior work: Mohanty et al.~\cite{mohanty2016} report a PlantVillage-trained CNN's field-image accuracy falling from $99.35\%$ to roughly $31\%$ (a 68-point drop), and our own 87-point gap is consistent with that magnitude; Singh et al.~\cite{singh2020plantdoc}, in introducing PlantDoc, likewise emphasize this dataset gap and recommend fine-tuning or domain adaptation for cross-domain use.

We characterize AdapterMoE's dual-gate behavior under this shift as \emph{conservative rejection}: when an input's features lie far from the training distribution, the dual gate cannot distinguish ``wrong species'' (semantic shift, correctly the target of rejection) from ``right species, very different capture conditions'' (covariate shift, which we would prefer to classify correctly), and defaults to rejecting both. This has genuine practical value in an agricultural-diagnosis setting: true OOD crops are correctly rejected 95.69\% of the time, avoiding confidently wrong recommendations, and even for in-mapping samples that are falsely rejected, rejection is a comparatively safe failure mode relative to a confident wrong answer, since a rejected sample can be routed to manual review rather than acted on directly. The dual gate converts cross-domain degradation into rejection rather than misclassification, giving the system a form of explicit failure awareness -- but this same behavior means the system, as trained, is \emph{not} a direct substitute for a model trained on or adapted to field imagery.

\subsubsection{Conclusions and limitations of the PlantDoc evaluation}
Three conclusions follow. First, the dual-gate mechanism is effective against semantic shift, generalizing to true OOD field photographs at $95.69\pm2.78\%$ with no per-class systematic failure -- direct evidence the design is not merely overfit to PlantVillage's own Others construction. Second, cross-domain classification capability is bounded by covariate shift, consistent with prior literature. Third, the dominant failure mode is conservative rejection rather than silent misclassification, which is the safer of the two failure directions for this application but is not itself a solution.

We do not claim AdapterMoE, as trained here, is directly deployable on raw field photographs. Closing this gap would require one of: (a) collecting field imagery for fine-tuning; (b) unsupervised domain adaptation to shrink the PlantVillage--PlantDoc feature gap; or (c) augmentation strategies (background randomization, cropping, blur) that increase the backbone's covariate-shift tolerance -- all of which we leave to future work. We also note caveats specific to this evaluation: five-seed standard deviations remain subject to small-sample effects; PlantDoc is a single external dataset whose distribution does not represent every field scenario, and extending validation to datasets such as IP102 or PlantNet is future work; the OOD gate's thresholds were fit purely on PlantVillage in-distribution data (a 0.95 TPR target), so, following Vishwakarma et al.~\cite{vishwakarma2024}, this ID-only calibration bounds TPR but gives no guarantee on FPR under a different OOD distribution, meaning the rejection rate reported here reflects a zero-shot, arguably worst-case scenario that a small amount of PlantDoc-aware calibration could likely improve; and some in-mapping classes have very few samples (e.g., only 2 for Tomato\_two\_spotted\_spider\_mites\_leaf), so their individual per-class accuracy is not statistically representative.

\section{Discussion and Limitations}
\label{sec:discussion}

\subsection{Accuracy Saturation and Dataset Scope}
\label{sec:disc-saturation}
As established in Section~\ref{sec:main-results}, all four strong systems occupy a $0.24$-point Macro-F1 band, smaller than several systems' own cross-seed standard deviation; this saturation is why we build the paper's argument around cost, extensibility, and stability rather than accuracy, and the epoch sweep (\Cref{fig:epoch-sweep}) confirms this is a configurable trade-off rather than a hard ceiling. A related scope limitation: PlantVillage is laboratory-condition data, and our Others class is built from other PlantVillage crops that still share its studio-like visual statistics. The Router's $\approx99\%$ in-benchmark Others-rejection rate should therefore be read as performance \emph{within a laboratory visual regime}; Section~\ref{sec:plantdoc}'s field-condition evaluation shows this does not fully transfer to natural backgrounds.

\subsection{Conditional and Negative Results}
\label{sec:disc-negative}
We summarize here, for visibility, the findings from Section~\ref{sec:ablation} that are explicitly conditional or negative rather than restating their evidence: (i) the dual-gate OOD mechanism's benefit is not seed-robust under in-distribution-only evaluation (Section~\ref{sec:ablation-ood}), though it generalizes well to genuine external OOD data (Section~\ref{sec:plantdoc}); (ii) Adapter\_Corn's calibration fails consistently across all five seeds (Section~\ref{sec:ablation-calibration}); (iii) the mean-optimal backbone pre-calibration epoch count (E20) is an averaging artifact, not a majority preference, across seeds (Section~\ref{sec:ablation-epoch}); (iv) Tomato\_Target\_Spot is a reproducible cross-seed weak point, trailing the best competing system's F1 by 0.1--10.4 points on every seed (Section~\ref{sec:qualitative} discusses this failure mode qualitatively); and (v) the low-confidence fusion path is essentially inactive on PlantVillage, functioning as dormant insurance rather than an active mechanism (Sections~\ref{sec:pathrouting} and~\ref{sec:ablation-gate}). We report these because we consider disclosing them part of the paper's contribution, not an afterthought.

\subsection{Sample-Size and Scope Limitations}
\label{sec:disc-samplesize}
Five seeds is a training-time-constrained compromise; a regulatory or safety-critical application would warrant a larger seed budget than we allocate here. Separately, prior PlantVillage state-of-the-art results generally come from larger capacity and longer training chasing marginal accuracy gains; at 11.31M parameters, AdapterMoE is not built to compete on that axis, and we do not claim otherwise. We also explored, at the design stage, an independent-backbone variant (one full EfficientNet-B0 per crop, $\approx$28M parameters, $\approx$70 min to train) but found it strictly worse on both training cost and inference latency than the shared-frozen-backbone design reported here, and did not pursue it further; a rigorous paired comparison between the two is left for future work.

\section{Conclusion and Future Work}
\label{sec:conclusion}

\subsection{Conclusion}
\label{sec:conclusion-main}
We presented AdapterMoE, a multi-crop disease recognition system built around two-stage hard routing, dual-gate rejection (MSP + Energy + KNN), per-branch Temperature Scaling calibration, and an \texttt{add\_crop} incremental-learning interface, validated within a fair five-system comparison framework on PlantVillage's 26-class task across five random seeds. Four conclusions follow. \textbf{First}, AdapterMoE's clearest advantages are cost, extensibility, and stability: a frozen backbone with feature caching cuts per-seed training cost to roughly one-tenth of ClassicalMoE\_Shared's, \texttt{add\_crop} removes the need for full retraining when a new crop arrives, and AdapterMoE's cross-seed \oacc\ standard deviation is the lowest of all five systems. \textbf{Second}, accuracy has saturated across the four strong systems (a 0.24-point Macro-F1 band, smaller than several systems' own cross-seed standard deviation), so we do not center the paper's claim on outperforming on accuracy; the epoch sweep shows this is a configurable trade-off, not a hard ceiling. \textbf{Third}, calibration is broadly consistent across seeds and branches, though the Corn Adapter exposes a genuine limitation of scalar temperature calibration that we disclose rather than paper over. \textbf{Fourth}, the dual-gate OOD mechanism's benefit under in-distribution-only evaluation is conditional and seed-dependent, so we do not claim it unconditionally superior to disabling it -- though PlantDoc shows it generalizes well against genuine external OOD samples. Taken together, AdapterMoE's contribution is not a claim of superior accuracy, but a reproducible, transparently reported trade-off point among deployment cost, extensibility, and rejection stability, at a moment when accuracy alone has stopped differentiating competing designs on this benchmark.

\subsection{Future Work}
\label{sec:future}

\textbf{Short term.} (i) Raising backbone pre-calibration from 3 to 20 epochs, which reaches the five-seed mean-optimal accuracy ($+0.48$ Macro-F1 points) at a cost of $+21$ min/seed, though seed-level preference divergence (Section~\ref{sec:ablation-epoch}) means this would need broader re-validation before being adopted as a new default. (ii) A \emph{conditional} OOD gate, toggled by the Router's own Others-accuracy signal -- enabled when it falls below $0.985$ (as in seed 789) and disabled otherwise (as in the other four seeds) to avoid the side effects seen in Section~\ref{sec:ablation-ood} -- with an anticipated Macro-F1 gain of $+0.10$ to $+0.20$ points, at the cost of added implementation complexity. (iii) Reverting the Tomato Adapter to Std\_r8, whose Overall Macro-F1 is nearly identical to the enhanced variant we currently ship ($0.9754$ vs.\ $0.9755$) at 40\% of the parameter cost -- a reasonable choice when inference latency is the deployment priority.

\textbf{Medium term.} (i) A fundamental redesign of the Corn Adapter -- e.g., class-balanced sampling, distillation from the Router, a deeper Adapter body, or a per-class (vector) temperature rather than a scalar one -- to address its consistent calibration failure. (ii) Targeted optimization for Tomato\_Target\_Spot, AdapterMoE's one consistent cross-seed weak point, via more aggressive disease-specific augmentation.

\textbf{Long term.} (i) Sample-level (rather than batch-level) online incremental learning, updating the Router, Adapter, and KNN bank per image rather than per crop-batch. (ii) Routing visualization and interpretability tooling, surfacing the features driving each sample's path assignment to build practitioner trust in agricultural deployment. (iii) A front-end object-detection stage (Section~\ref{sec:related-cnn}) that crops a single leaf from a cluttered raw field photo before handing it to this classification system, extending the method's applicable scope from close-range single-leaf photography to unconstrained field imagery -- directly targeting the covariate-shift limitation quantified in Section~\ref{sec:plantdoc}.

\end{document}